\documentclass[twoside,11pt]{article}
\usepackage{jair, theapa, rawfonts}
\usepackage{algorithm,algpseudocode}
\usepackage{amsthm}
\usepackage{amsfonts}
\usepackage{amssymb}
\usepackage{amsmath}
\usepackage{balance}
\usepackage{courier}
\usepackage{color}
\usepackage{float}
\usepackage{graphicx}
\usepackage[space]{grffile}
\usepackage{helvet}
\usepackage{subcaption}
\usepackage{float}
\usepackage{multirow}
\usepackage{mathrsfs}
\usepackage{tabularx}
\usepackage{times}
\usepackage{tensor}
\usepackage{rotating}
\usepackage{url}
\usepackage{xfrac}

\DeclareMathOperator*{\argmin}{arg\!min}
\DeclareMathOperator*{\argmax}{arg\!max}
\newcommand{\tuple}[1]{\ensuremath{\left \langle #1 \right \rangle }}

\newcolumntype{C}[1]{>{\centering\let\newline\\\arraybackslash\hspace{0pt}}m{#1}}

\graphicspath{ {figs/} }

\ShortHeadings{Human-Machine Collaborative Optimization}
{Gombolay, Jensen, Stigile, Golen, Shah, \& Shah}
\firstpageno{1}

\begin{document}

\title{Human-Machine Collaborative Optimization \\via Apprenticeship Scheduling}

\author{\name Matthew Gombolay \email gombolay@mit.edu \\
\addr Massachusetts Institute of Technology, \\
77 Massachusetts Avenue,\\
Cambridge, MA 02114 USA
\AND
\name Reed Jensen \email rjensen@ll.mit.edu \\
\name Jessica Stigile \email jessica.stigile@ll.mit.edu \\
\addr MIT Lincoln Laboratory,\\
244 Wood Street, \\
Lexington, MA 02420 USA \\
\AND
\name Toni Golen \email tgolen@bidmc.harvard.edu \\
\name Neel Shah \email ntshah@bidmc.harvard.edu \\
\addr Beth Israel Deaconess Medical Center, \\
330 Brookline Avenue, \\
Boston, MA 02215 USA \\
\AND 
\name Sung-Hyun Son \email sson@ll.mit.edu \\
\addr MIT Lincoln Laboratory,\\
244 Wood Street, \\
Lexington, MA 02420 USA \\
\AND
\name Julie Shah \email julie\_a\_shah@csail.mit.edu \\
\addr Massachusetts Institute of Technology,\\
77 Massachusetts Avenue,\\
Cambridge, MA 02114 USA}


\maketitle

\begin{abstract}
\begin{quote}
	
	Coordinating agents to complete a set of tasks with intercoupled temporal and resource constraints is computationally challenging, yet human domain experts can solve these difficult scheduling problems using paradigms learned through years of apprenticeship. A process for manually codifying this domain knowledge within a computational framework is necessary to scale beyond the ``single-expert, single-trainee" apprenticeship model. However, human domain experts often have difficulty describing their decision-making processes, causing the codification of this knowledge to become laborious. We propose a new approach for capturing domain-expert heuristics through a pairwise ranking formulation. Our approach is model-free and does not require enumerating or iterating through a large state space. We empirically demonstrate that this approach accurately learns multifaceted heuristics on a synthetic data set incorporating job-shop scheduling and vehicle routing problems, as well as on two real-world data sets consisting of demonstrations of experts solving a weapon-to-target assignment problem and a hospital resource allocation problem. We also demonstrate that policies learned from human scheduling demonstration via apprenticeship learning can substantially improve the efficiency of a branch-and-bound search for an optimal schedule. We employ this human-machine collaborative optimization technique on a variant of the weapon-to-target assignment problem. We demonstrate that this technique generates solutions substantially superior to those produced by human domain experts at a rate up to $9.5$ times faster than an optimization approach and can be applied to optimally solve problems twice as complex as those solved by a human demonstrator.
	
\end{quote}
\end{abstract}

\section{Introduction}

Resource scheduling and optimization is a costly, challenging problem that affects almost every aspect of our lives. In healthcare, for example, patients with non-urgent needs who experience prolonged wait times have higher rates of treatment noncompliance and missed appointments \cite{Kehle:2011,Pizer:2011}. In military engagements, the weapon-to-target assignment problem requires warfighters to deploy the minimal amount of resources in order to mitigate as many threats as possible while maximizing the duration of survival \cite{Lee:2003}. 

The problem of optimal task allocation and sequencing with upper- and lowerbound temporal constraints (i.e., deadlines and wait constraints) is NP-Hard~\cite{Bertsimas:2005}, and domain-independent approaches to real-world scheduling problems quickly become computationally intractable~\cite{boese1994new,streeter2006landscape,do2003sapa}. However, human domain experts are able to learn from experience to develop strategies, heuristics and rules-of-thumb to effectively respond to these problems. The challenge we pose is to autonomously learn the strategies employed by these domain experts; this knowledge can be applied and disseminated more efficiently with such a model than with a ``single-expert, single-apprentice" model.

Researchers have made significant progress toward capturing domain-expert knowledge from demonstration \cite{Berry:2011,Abbeel:2004,Konidaris:2011,Zheng:2014,Odom:2015,Vogel:2012,Ziebart:2008}. In one recent work \cite{Berry:2011}, an AI scheduling assistant called PTIME learned how users preferred to schedule events. PTIME was subsequently able to propose scheduling changes when new events occurred by solving an integer program. Two limitations to this work exist, however: PTIME requires users to explicitly rank their preferences about scheduling options to initialize the system, and also uses a complete solver that, in the worst-case scenario, must consider an exponential number of options. 

Research focused on capturing domain knowledge based solely on user demonstration has led to the development of inverse reinforcement learning (IRL) \cite{Abbeel:2004,Konidaris:2011,Zheng:2014,Odom:2015,Vogel:2012,Ziebart:2008}. IRL serves the dual purpose of learning an unknown reward function for a given problem and learning a policy to optimize that reward function. 

However, there are two primary drawbacks to IRL for scheduling problems, computational tractability and the need for an environment model. The classical apprenticeship learning algorithm, developed by Abbeel and Ng in 2004, requires repeated solving of a Markov decision process (MDP) until a convergence criterion is satisfied. However, enumerating a large state space, such as those common to large-scale scheduling problems involving hundreds of tasks and tens of agents, can quickly become computationally intractable due to memory limitations. Approximate dynamic programming approaches exist that essentially reformulate the problem as regression \cite{Konidaris:2011,Mnih:2015}, but the amount of data required to regress over a large state space remains challenging, and MDP-based scheduling solutions exist only for simple problems \cite{Wu:2011,Wang:2005,Zhang:1995}.

IRL also requires a model of the environment for training. At its most basic, reinforcement learning uses a Markovian transition matrix that describes the probability of transitioning from an initial state to a subsequent state when taking a given action. In order to address circumstances in which environmental dynamics are unknown or difficult to model within the constraints of a transition, researchers have developed Q-Learning and its variants, which have had much recent success \cite{Mnih:2015}. However, these approaches require the ability to ``practice," or explore the state space by querying a black-box emulator to solicit information about how taking a given action in a specific state will change that state. 

Another prior method involves directly learning a function that maps states to actions \cite{Chernova:2007,Terrell:2012,Huang:2014}. For example, Ramanujam and Balakrishnan trained a discrete-choice model using real data collected from air traffic controllers, and showed how this model can accurately predict the correct runway configuration for an airport \cite{Ramanujam:2011}. Sammut et al. \cite{Sammut:1992} applied a decision tree model for an autopilot to learn to control an aircraft from expert demonstration. Action-driven learning techniques offer great promise for learning policies from expert demonstrators, but they have not been applied to complex scheduling problems. However, in order for these methods to succeed, the scheduling problem must be modeled in a way that allows for efficient computation of a scheduling policy.

In this paper, we propose a technique, which we call ``apprenticeship sch eduling," to capture this domain knowledge in the form of a scheduling policy. Our objective is to learn scheduling policies through expert demonstration and validate that schedules produced by these policies are of comparable quality to those generated by human or synthetic experts. Our approach efficiently utilizes domain-expert demonstrations without the need to train with an environment emulator. Rather than explicitly modeling a reward function and relying upon dynamic programming or constraint solvers -- which become computationally intractable for large-scale problems of interest -- our objective is to use action-driven learning to extract the strategies of domain experts in order to efficiently schedule tasks. 

The key to our approach is the use of pairwise comparisons between the actions taken (e.g., schedule agent $a$ to complete task $\tau_i$ at time $t$) and the set of actions not taken (e.g., unscheduled tasks at time $t$) to learn the relevant model parameters and scheduling policies demonstrated by the training examples.  
\color{black} Our approach was inspired by cognitive studies of human decision-making, in which learning through comparisons -- and, in particular, paired comparisons -- was identified as a foundation of human multi-criteria decision-making \color{black}~\cite{saaty2008relative,lombrozo2006structure}\color{black}. 
Rather than explicitly query human experts about their preferences, our approach functions more like a human apprentice who learns by observing a sequence of actions performed by a demonstrator. Our approach automatically computes\color{black}~pairwise comparisons of the features describing the action taken at each moment in time relative to the corresponding set of actions not taken, producing sets of both positive and negative training examples. We formulate the apprenticeship scheduling problem as one of learning a pairwise preference model, and construct a classifier that is able to predict the rank of all possible actions and, in turn, predict which action the expert would ultimately take at each moment in time.

We validated our approach using both a synthetic data set of solutions for a variety of scheduling problems and two real-world data sets of demonstrations by human experts solving a variant of the weapon-to-target assignment problem \cite{Lee:2003}, known as anti-ship missile defense (ASMD), and a hospital resource allocation problem~\cite{Gombolay:2016b}. The synthetic and real-world problem domains we used to empirically validate our approach represent two of the most challenging classes within the taxonomy established by \color{black}\citeA{Korsah:2013}\color{black}. 

The first problem we considered was the vehicle routing problem with time windows, temporal dependencies and resource constraints (VRPTW-TDR). Depending upon parameter selection, this family of problems encompasses the traveling salesman (Type 1), job-shop scheduling, multi-vehicle routing and multi-robot task allocation problems, among others. We found that apprenticeship scheduling accurately learns multifaceted heuristics that emulate the demonstrations of experts solving these problems. We observed that an apprenticeship scheduler trained on a small data set of 15 scheduling demonstrations selected the correct scheduling action with up to $95\%$ accuracy. \color{black} We also empirically characterized the extent to which our method is robust to errors that humans -- even experts -- may commonly make. We found that our method is able to learn a high-quality representation of the demonstrator's underlying heuristic from a ``noisy" expert demonstrator that selects an incorrect action up to $20\%$ of the time. \color{black}

Next, we observed that apprenticeship scheduling learned a policy for ASMD that outperformed the average ASMD domain expert for a statistically significant portion of problem scenarios ($p < 0.05$) when trained on 15 perfect expert-generated schedules. Third, we trained a decision support tool to assist nurses in managing resources -- including patient rooms, staff and equipment -- in a Boston hospital. We found that $90\%$ of the high-quality recommendations generated by the apprentice scheduler were accepted by the nurses and doctors participating in the study. 

In this work, we also introduce a new technique called Collaborative Optimization via Apprenticeship Scheduling (COVAS), which incorporates learning from human expert demonstration within an optimization framework to automatically and efficiently produce optimal solutions for challenging real-world scheduling problems. This technique applies apprenticeship scheduling to generate a favorable (if suboptimal) initial solution to a new scheduling problem. To guarantee that the generated schedule is serviceable, we augment the apprenticeship scheduler to solve a constraint satisfaction problem, ensuring that the execution of each scheduling commitment does not directly result in infeasibility for the new problem. COVAS uses this initial solution to provide a tight bound on the value of the optimal solution, substantially improving the efficiency of a branch-and-bound search for an optimal schedule.

We first presented the apprenticeship scheduling technique in a prior work~\cite{Gombolay:2016a}, and also previously discussed an application of the technique to the hospital scheduling problem~\cite{Gombolay:2016b}. This paper incorporates multiple extensions to these original works: First, we improve the performance of the original technique through the use of hyperparamter tuning. Second, we incorporate the data set acquired from the hospital domain in the previous study \cite{Gombolay:2016b} to validate apprenticeship scheduling using a second real-world data set consisting of scheduling decisions generated by hospital nurses. Third, we present COVAS, an algorithmic extension that enables human-machine collaborative optimization. COVAS leverages apprenticeship scheduling to optimally solve scheduling problems, whereas apprenticeship scheduling alone does not provide guarantees for solution quality. We report here that COVAS is able to leverage viable (but imperfect) human demonstrations to quickly produce globally optimal solutions. Fourth, we show that COVAS can transfer an apprenticeship scheduling policy learned for a small problem to optimally solve problems involving twice as many variables as those observed during any training demonstrations, and also produce an optimal solution an order of magnitude faster than mathematical optimization alone. 


\section{Background}

In this section, we briefly review goal and policy learning, as well as methods for bridging machine learning (ML) and optimization. We also discuss the applicability and limitations of prior works related to learning through scheduling demonstration. 

\subsection{Goal Learning}
Here, we review both IRL-based techniques and methods proposed for recommender and preference-learning systems within the realm of goal learning.
\subsubsection{Inverse Reinforcement Learning}
\label{sec:LfD}
Learning from demonstration (LfD) is an active subfield of ML \cite{Abbeel:2004,Berry:2011,Ijspeert:2002,Konidaris:2011,Zheng:2014,Odom:2015,Terrell:2012,Thomaz:2006,Vogel:2012,Ziebart:2008}. Arguably, the most ubiquitous approach to LfD is inverse reinforcement learning, which is founded on a Markov decision process $M = (S,A,T,\gamma,R)$ where:

\begin{itemize}
\item{S is a set of states.}
\item{A is a set of actions.}
\item{$T: S \times A \times S \rightarrow [0,1]$ is a transition function, where $T(s,a,s')$ is the probability of being in state $s'$ after executing action $a$ in state $s$.}
\item{$R$: $S \rightarrow \mathbb{R}$ ($S \times A \rightarrow \mathbb{R}$) is a reward function that takes the form of $R(s)$ or $R(s,a)$ depending upon whether the reward is assessed for being in a state or for taking a particular action within a state.}
\item{$\gamma \in [0,1)$ is the discount factor for future rewards.}
\end{itemize}

In a Markov decision process, the goal is to learn a policy $\pi: S \rightarrow A$ that dictates which action to take in each state in order to maximize the infinite-horizon expected reward starting in state $s$. This reward is defined by a value function, $V^{\pi}(s)$, as shown in Equation \ref{eq:valueFunction}:

\begin{equation}
V^{\pi}(s) = \mathbb{E}_{\pi} \left[\sum_{t=0}^T \gamma^t R(s_t) | s_o = s\right]
\label{eq:valueFunction}
\end{equation}
The value function satisfies the Bellman equation for all $s \in S$, as shown in Equation \ref{eq:BellmanEquation}.
\begin{equation}
V^{\pi}(s) = R(s) + \gamma \left[\sum_{s'\in S} T(s,\pi(s),s')V^{\pi}(s')\right]
\label{eq:BellmanEquation}
\end{equation}
A policy $\pi$ is an optimal policy $\pi^*$ \emph{iff} $\forall s\in S$ Equation \ref{eq:OptimalPi} holds.
\begin{equation}
\pi(s) = \argmax_{a\in A} \left( \sum_{s' \in S} T(s,a,s')\left(R(s')+\gamma V^{\pi}(s')\right) \right)
\label{eq:OptimalPi}
\end{equation}

The problem of inverse reinforcement learning (IRL) is to take as input 1) a Markov decision process without a known reward function $R$ and 2) a set of $m$ expert demonstrations $O = \{(s_o,a_o),(s_1,a_1),\ldots,(s_m,a_m)\}$, and to then determine a reward function $R$ that produces the expert demonstrations. IRL has previously been successfully applied to autonomous driving~\cite{Abbeel:2004}, aerobatic helicopter flight~\cite{abbeel2007application}, urban navigation~\cite{Ziebart:2008}, spoken dialog systems~\cite{chandramohan2011user}, and more. Researchers have also extended the capability of IRL algorithms to enable learning from operators with differing skill levels \cite{Ramachandran:2007} and identification of operator subgoals \cite{Michini:2012}.

The computational bottleneck of IRL and dynamic programming, in general, is the size of the state space. Algorithms that solve the IRL problem~\cite{lagoudakis2003least,Sutton:1999,tesauro1995temporal,watkins1992q} typically work by iteratively updating the estimate of the future expected reward of each state until convergence. However, for many problems of interest, the number of states is too numerous to hold in the memory of modern computers, and the time required for the expected future reward to converge can be impractical~\cite{Wu:2011,Wang:2005,Zhang:1995}. 

Even if one approximately solves the RL problem \cite{Konidaris:2011,Sutton:1999}, RL is still ill-suited for handling the temporal dependencies among tasks inherent in scheduling problems. Some researchers have attempted to extend the traditional Markov decision process to characterize temporal phenomena, but these techniques do not scale efficiently \cite{Bradtke:1994,Das:1999,Yu:2009}. The inherent challenge is that complex real-world scheduling problems are highly non-Markovian: the next state of the environment is dependent upon the history of actions taken to arrive at the current state and time. The few works that have addressed scheduling problems via RL assume models that are too restrictive: tasks must be periodic, occur with a regular frequency, and be independent, meaning there are no temporal dependencies between the tasks~\cite{Zhang:1995,Wu:2011}. Even work~\cite{aydin2000dynamic} that relaxes the assumption of determinism and allows for tasks comprising predefined subtasks linked through precedence (as opposed to tasks representing atomic units of work) still does not consider wait-, deadline-, or resource-based constraints, nor does it consider problems in the \textbf{XD} complexity class~\cite{Korsah:2013}.

\subsubsection{Recommender/Preference-Learning Systems}
While not typically considered LfD, recommender systems are important within the field of goal learning. Recommender systems -- those that use collected information to predict a rating or degree of preference a consumer would give for an item (e.g., goods or services) --  have become ubiquitous during the Internet age, including services such as Netflix, which predicts which movies a viewer would want to watch~\cite{Koren:2009}. These systems generally fall into one of two categories: collaborative filtering (CF) or content-based filtering (CB) \cite{Park:2012}. In essence, collaborative filtering is a technique through which an algorithm learns to predict content for a single user based upon his or her history and that of other users who share his or her interests. However, CF suffers from problems related to  data sparsity and scalability \cite{Park:2012}. CB works by comparing content that the user has previously viewed with new content \cite{Claypool:1999,Herlocker:2004,Sarwar:2000}. The challenge of content-based filtering lies in the difficulty of measuring the similarities between two items; also, these systems can often over-fit, only predicting content that is very similar to that which the user has previously used \cite{Basu:1998,Schafer:2007}. Researchers have previously employed association rules \cite{Cho:2002}, clustering \cite{Lihua:2005,Berry:2004}, decision trees \cite{Kim:2002}, k-nearest neighbor algorithms \cite{Kim:2009}, neural networks \cite{Anders:1999,Ibnkahla:2000}, link analysis \cite{Cai:2004}, regression \cite{Malhotra:2010}, and general heuristic techniques \cite{Park:2012} to recommend content to users.

Ranking the relevance of Web pages is a key focus within systems that recommend suggested topics to users \cite{Cao:2007,Haveliwala:2002,Herbrich:2000,Jin:2008,Page:1999,Pahikkala:2007,Platt:2007,Valizadegan:2009,Volkovs:2009}. The seminal paper on Web page ranking by Page et al. initiated the computational study of page ranking with an algorithm, PageRank, which assesses the relevance of a page by determining the number of other pages that link to the page in question \cite{Page:1999}. Since that paper, many have focused on developing better models for recommending Web pages to users; these models can then be trained using various ML algorithms \cite{Haveliwala:2002,Herbrich:2000,Jin:2008,Pahikkala:2007}. 

There are three primary approaches to modeling the importance of a Web page: pointwise, pairwise, and listwise ranking. In pointwise ranking, the goal is to determine a score for a Web page via regression analysis, given features describing its contents \cite{Platt:2007,Page:1999}. Pairwise ranking is typically a classification problem in which the aim is to predict whether one page is more relevant than another, given a user's query \cite{Jin:2008,Pahikkala:2007}. More recent efforts have focused on listwise ranking, in which researchers develop loss-functions based on entire lists of ranked Web pages, rather than individual pages or pairwise comparisons between pages \cite{Cao:2007,Valizadegan:2009,Volkovs:2009}. Our approach draws inspiration from the Web page pairwise ranking formulation in order to improve the tractability of learning scheduling policies from demonstration. We further discuss the relationship between prior work and our own approach in Section \ref{sec:TechnicalApproach}.

The recommender and preference-learning system most closely related to ours is that of Berry et al. \cite{Berry:2006,Berry:2011}, which focused specifically on scheduling applications. Their goal was to develop an autonomous scheduling assistant that learned the preferences of the user. Berry et al. produced a number of works over the course of a decade, culminating in the development of an automated scheduling assistant, called PTIME. The purpose of PTIME was to help human coworkers schedule meetings. Berry et al. incorporated extensive questionnaires to solicit the preferences of human workers regarding how they preferred to arrange their schedules. PTIME would take these preferences as input and map them to a mathematical objective function. When a new meeting needed to be arranged amongst the workers, PTIME would solve a mixed-integer mathematical program to determine the optimal time for this meeting to occur. However, after approximately a decade of work, the ultimate acceptance rate of PTIME's suggestions was only $60\%$. These authors conducted a retrospective analysis of their work and presented the following guidance for future researchers \cite{Berry:2011}:
\begin{enumerate}
\item{``A personal assistant must build trust."}
\item{``An assistive agent must aim to support, rather than replace, the user's natural process."}
\end{enumerate}
These tenants have served as an inspiration for our own work, and we believe all future works should begin with these key design principles.

Other works have outlined alternate approaches to elicitation and utilization of user preferences. De Grano et al. presented a method for optimizing scheduling shifts among nurses by soliciting nurses' preferences via an auction process \cite{Grano:2009}. In particular, De Grano et al. used an iterative approach in which nurses first bid on which shifts they would prefer; then, their algorithm matches nurses to shifts based on their collective bids. Next, the nurses view the results and adjust their bids to push the algorithm toward a more preferable result. This process repeats over a number of iterations. The need for this iterative approach is due to the fact that nurses' preferences were not independent: each nurse's preferences would change according to the preferences of others. Further, it was not feasible for De Grano et al. to codify a rule set or learn a policy for each nurse~\cite{Grano:2009}.

Boutilier et al. and others \cite{Boutilier:2004,Boutilier:1999,Ozturke:2005} alternatively focused on modeling preferences as a set of \emph{ceteris paribus} (all other things being equal) preference statements. In these works, researchers solicited preferences from users, typically in the form of binary comparisons. For example, consider the problem of determining which food and drink to serve a guest \cite{Boutilier:2004}. In this scenario, one may already know the following:
\begin{itemize}
\item{The guest prefers to drink red over white wine when eating a steak.}
\item{The guest prefers steak over chicken.}
\item{The guest prefers to drink white wine when eating chicken.}
\end{itemize}
Determining the optimal food/drink pairing can be performed in polynomial-time; however, identifying the relative optimality two pairings is NP-complete \cite{Boutilier:2004}.

Other researchers have focused on developing techniques for efficiently incorporating preferences into constraint satisfaction problems \cite{dubois1999computing,lin2005solving,rossi2009preferences,rudova2002university,schiex1995valued,Soomer:2008}. A subset of this work has specifically addressed the unique challenges of solving such formulations for scheduling problems \cite{benton2012temporal,khatib2001temporal,minton1992minimizing,morris2004strategies,peintner2004low,yorke2003temporal,rossi2006uncertainty}.


These methods, which are designed for scheduling problems, still suffer from issues with computational tractability. As mentioned previously, Berry et al. used a preference learning algorithm to codify an objective function, which could then be solved via mathematical optimization \cite{Berry:2006}. Similarly, Wilcox et al. used mathematical programming to maximize the incorporation of users' scheduling preferences into the system \cite{Wilcox:2012}. However, mathematical programming is not a tractable solution technique for many real-world scheduling problems \cite{Bertsimas:2005}, including the anti-ship missile defense and hospital resource allocation problems presented in this work. Solving these problems typically requires specification of domain-specific heuristics in order to focus the search space. In this work, we present a system designed to automatically learn a heuristic policy from expert demonstration, and then apply the heuristic in order to intelligently explore the search space, reducing computation time. 

\subsection{Policy Learning}
One alternative approach to goal learning is policy learning, which focuses on learning a mapping from states to actions \cite{Chernova:2007,Huang:2014,Sammut:1992,Ramanujam:2011}. This technique has been applied to learn cognitive decision-making tasks from human experts \cite{Ramanujam:2011,Sammut:1992,Silver:2016,Inamura:1999,Rybski:1999}, including an air traffic control task \cite{Ramanujam:2011} and a piloting task \cite{Sammut:1992}. 

Ramanujam and Balakrishnan investigated learning a discrete-choice model for how air traffic controllers decide which runways to use for arriving and departing aircraft according to weather, arrival and departure demand, and other environmental factors. The authors trained a discrete-choice model on real data from air traffic controllers and showed how the model was able to accurately predict the correct runway configuration for the airport \cite{Ramanujam:2011}. 

Sammut et al. applied a decision tree model to train an airplane's autopilot from expert demonstration. Their approach generates a separate decision tree for each of the following control inputs: elevators, ailerons, flaps, and thrust. In their investigation, Sammut et al. noted that each pilot demonstrator could execute a planned flight path differently. These demonstrations could be in disagreement, thus making the learning problem significantly more difficult. To cope with the variance between pilot executions, the system learned a separate model for each pilot \cite{Sammut:1992}. 


Other systems learn policies through interaction and feedback, as well as demonstration, from the user \cite{baranes2013active,bullard2016grounding,Chernova:2008,grollman2008sparse,Inamura:1999,konidaris2011robot,zeng2016learning}. 
For example, Chernova and Veloso developed a Gaussian mixture model able to interactively learn from demonstration \cite{Chernova:2007}. Their algorithm first learns a reasonable policy for a given task (e.g., driving a car along a highway), then solicits user feedback by constructing scenarios involving a high level of uncertainty. Support vector machines are then applied to learn when an autonomous agent should request additional demonstrations \cite{Chernova:2008}. 

Policy learning is an important complement to goal- or reward-learning. While goal- and reward-learning approaches are able to capture high-level goals in order to produce quality schedules \cite{Abbeel:2004,Berry:2006}, these methods are limited by their reliance on computational methods for exploring the search space to identify a high-quality schedule. IRL relies on dynamic programming, which requires state space enumeration, while approaches such as PTIME \cite{Berry:2006} rely upon mathematical programming. Policy learning, on the other hand, is well-suited to guiding exploration of a state space. With a function mapping states to actions, a system can construct a schedule by taking sequential scheduling actions (e.g., assigning a worker to a task at the present time). In this sense, a learned policy can serve as a type of domain-specific heuristic to intelligently guide a search within a large state space. However, we are unaware of any prior attempts to apply policy learning to the scheduling domain. 

\subsection{Blending Machine Learning and Optimization}
\label{sec:COVASBackground}
Typically, reward and policy learning are limited by the quality of the relevant demonstrations. However, even if the demonstrations are high-quality, one cannot assume demonstrators nor their demonstrations will be optimal -- or even uniformly suboptimal \cite{Aleotti:2006,Sammut:1992}. As such, some have sought to directly model the sub-optimality of demonstrations. For example, Zheng et al. cleverly extended the work of Ramachandran and Amir \cite{Ramachandran:2007} to model the trustworthiness of the demonstrator within a softmax formulation transition function for reinforcement learning \cite{Zheng:2014}, as shown in Equation \ref{eq:BIRL2}. In this equation, $Q^{\pi^*(R)}(s,a)$ is the expected reward for taking action $a$ in state $s$, assuming reward function $\boldsymbol{R}$ with the associated optimal policy $\pi^*$:
\begin{equation}
Pr((s,a)|\alpha;\boldsymbol{R}) = \frac{e^{\alpha Q^{\pi^*(\boldsymbol{R})}(s,a)}}{\sum_{a'}e^{\alpha Q^{\pi^*(\boldsymbol{R})}(s,a')}}
\label{eq:BIRL2}
\end{equation}
Through such a mechanism, it is possible to learn a policy that outperforms human demonstrators by inferring the intended goal rather than the demonstrated goal. Zheng et al. showed that their approach was better able to capture the ground-truth objective function from imperfect training data than Bayesian IRL~\cite{Ramachandran:2007}, which does not include a trustworthiness parameter for demonstrations. They validated their approach using a synthetic data set in an experiment with the goal of identifying the best route through an urban domain. However, one limiting assumption from their work is that a system is able to accurately measure the trustworthiness of the demonstrations -- especially the relative trustworthiness amongst the demonstrations.

AlphaGo is another well-known ML-optimization framework recently developed to play Go, a turn-based strategy game \cite{Silver:2016}. At its core, AlphaGo is based on policy learning; it uses a Monte-Carlo Tree Search (MCTS) that is guided by a neural network policy trained on a data set of 30 million examples of demonstrations by human Go experts. A policy $\pi$ is employed to initially explore the search tree, and two additional components are used to evaluate the quality of each branching point in the tree. The first component is a second policy, $\pi'$, which is identical to the first except that the neural network includes fewer nodes. This smaller size enables the second policy to rapidly play the Go game to completion in order to predict a winner \cite{Silver:2016}. 

The second component of AlphaGo is a value function trained via Q-learning. The developers rewired and duplicated the initial policy $\pi$ to enable improvement through self-play. These duplicated, rewired policies $\pi_{Self-Play}$ would repeatedly play Go against one another and use a policy gradient approach, developed by Sutton et al., to iteratively improve their policies; the developers then captured a data set of 30 million moves taken by these policies\cite{Sutton:1999}. They then used this data set to train a Q-learning algorithm to predict the expected value of taking a given action in a given state. Interestingly, the authors noted that these self-play policies actually performed worse than the original $\pi$ trained on actual human demonstrations, but did not have a cohesive theory for why this was the case. Nonetheless, AlphaGo serves as a key example for how policy learning, coupled with optimization techniques (e.g., Q-learning and policy gradient methods) can yield performance on strategy games that is superior to that of humans.

The learning-optimization system most related to our work is that developed by Banerjee et al., who considered a scheduling problem for aircraft carrier flight deck operations. The system repeatedly solved a scheduling problem wherein the variables remained the same (i.e., variables describing which workers performed which tasks and when), but the constraints relating the variables (e.g. temporal constraints between tasks) changed \cite{Banerjee:2011}. Using a mixed-integer linear program (MILP) formulation, they proposed a ML-optimization pipeline in which the system performed a branch-and-bound search over the integer variables, and used the prediction of a regression algorithm trained on examples of previously solved problems to provide a provable lowerbound for the optimality of the current integer variable assignments. This approach relied upon the generation of a large database of solutions to train the regression algorithm; however, this generation requires the costly exercise of repeatedly solving a large set of MILPs, which can be intractable for large-scale scheduling problems.

\section{Model for Apprenticeship Learning}
\label{sec:apprenticeshipScheduling}
In this section, we present a framework for learning, via expert demonstration, a scheduling policy that correctly determines which task to schedule as a function of task state.

\subsection{Problem Domain}
We intend for our apprenticeship learning model to address a variety of scheduling problem types. Korsah et al. provided a comprehensive taxonomy for classes of scheduling problems, which vary according to formulation of constraints, variables and objective or utility function \cite{Korsah:2013}. Within this taxonomy, there are four classes addressing interrelated utilities and constraints: No Dependencies (ND \cite{LiuRSS:2013}), In-Schedule Dependencies (ID \cite{Brunet08_GNC,GombolayJAIS:2014,Nunes:2015}, Cross-Schedule Dependencies (XD \cite{GombolayRSS:2013}) and Complex Dependencies (CD \cite{Jones:2011}).

The Korsah et al. taxonomy also delineates between tasks requiring one agent (`single-agent tasks" [SA]); and tasks requiring multiple agents (``multi-agent tasks" [MA]). Similarly, agents that perform one task at a time are ``single-task agents" (ST), while agents capable of performing multiple tasks simultaneously are ``multi-task agents" (MT). Lastly, the taxonomy distinguishes between ``instantaneous assignment" (IA), in which all task and schedule commitments are made immediately, and ``time-extended assignment" (TA), in which current and future commitments are planned.

In this work, we demonstrate our approach for two of the most difficult classes of scheduling problems defined within this taxonomy: \textbf{XD [ST-SA-TA]} and \textbf{CD [MT-MA-TA]}. The first problem we consider is the VRPTW-TDR, which is an \textbf{XD [ST-SA-TA]}-class problem. We next consider two real-world problems within the more-difficult \textbf{CD [MT-MA-TA]} class. The second problem (first real-world domain) is a variant of the weapon-to-target assignment problem (WTA) \cite{Lee:2003}, known as anti-ship missile defense (ASMD). The third problem (second real-world problem) we address is one of hospital resource allocation on a labor and delivery unit, wherein one nurse, called the ``resource nurse," is responsible for ensuring that the correct patient is in the correct type of room at the correct time, with the correct types of nurses present to care for those patients. \color{black} The characteristics of the three problem domains we explore in evaluating the apprenticeship scheduling algorithm are shown in Table \ref{tab:myTable}.~\color{black}

\begin{table}
	\begin{center}
		\begin{tabular}{ l||c|c|c } 
			& \begin{minipage}{.2\textwidth}
				\includegraphics[width=\linewidth]{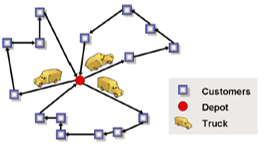}
			\end{minipage} & \begin{minipage}{.2\textwidth}
				\includegraphics[width=\linewidth]{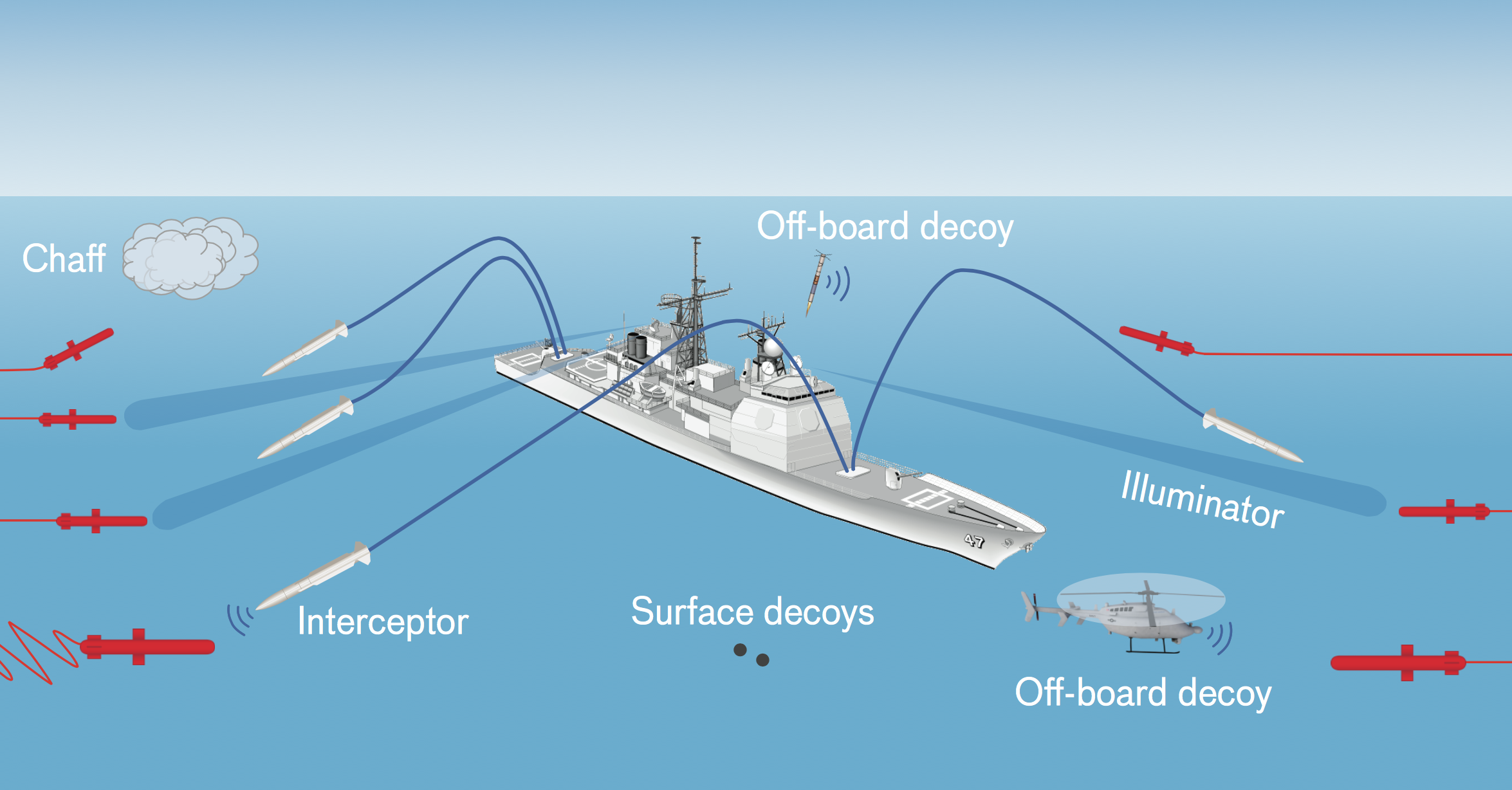}
			\end{minipage} & \begin{minipage}{.2\textwidth}
				\includegraphics[width=\linewidth]{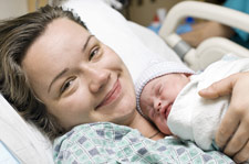}
			\end{minipage} \\ 
			Problem Domain & VRPTW-TDR & ASMD &  Hospital Resource Mngmt.\\ \hline\hline
			Describing Section & Section \ref{sec:DataSet_VRP} & Section \ref{sec:ASMDDataSet} & Section \ref{sec:LaborAndDeliveryDataSet} \\ \hline
			Data Type & Synthetic & Real-world & Real-world \\ \hline
			Dependency Type & XD & CD & CD\\ \hline
			Agent Type & ST & MT & MT\\ \hline
			Task Type & SA & MA & MA\\ \hline
			Allocation Type & TA & TA & TA\\
			\hline
		\end{tabular}
	\end{center}
	\caption{\color{black}This table summarizes the differing characteristics of the three problem domains used to empirically evaluate the apprenticeship scheduling algorithm.\color{black}}
	\label{tab:myTable}
\end{table}

\subsection{Technical Approach}
\label{sec:TechnicalApproach}
Many approaches to learning via demonstration, e.g., IRL, are based on Markov models \cite{Busoniu:2008,Barto:2003,Konidaris:2007,Puterman:2014}. Markov models, however, do not capture the temporal dependencies between states and are computationally intractable for large problem sizes. In order to determine which tasks to schedule at which times, we draw inspiration from the domain of Web page ranking \cite{Page:1999}, or predicting the most relevant Web page in response to a search query. One important component of page ranking is capturing how pages relate to one another as a graph with nodes (representing Web pages) and directed arcs (representing links between those pages) \cite{Page:1999}. This connectivity is a suitable analogy for the complex temporal dependencies (precedence, wait and deadline constraints) relating tasks within a scheduling problem.

Recent approaches to page ranking have focused on pairwise and listwise models, which each have advantages over pointwise models \cite{Valizadegan:2009}. In listwise ranking, the goal is to generate a ranked list of Web pages directly \cite{Cao:2007,Valizadegan:2009,Volkovs:2009}, while a pairwise approach determines ranking based on pairwise comparisons between individual pages \cite{Jin:2008,Pahikkala:2007}. We chose the pairwise formulation to model the problem of predicting the best task to schedule at time $t$. 

The pairwise model has key advantages over the listwise approach: First, classification algorithms (e.g., support vector machines) can be directly applied \cite{Cao:2007}. Second, a pairwise approach is non-parametric, in that the cardinality of the input vector is not dependent upon the number of tasks (or actions) that can be performed at any instance. Third, training examples of pairwise comparisons in the data can be readily solicited. From a given observation during which a task was scheduled, we only know which task was most important, not the relative importance between all tasks. Thus, we create training examples based on pairwise comparisons between scheduled and unscheduled tasks. A pairwise approach is more natural because we lack the necessary context to determine the relative rank between two unscheduled tasks.


\color{black} We formulate the apprenticeship scheduling problem as one of learning a pairwise preference model, as follows. Consider a set of $m$ observations, $O=\{O_1,O_2,\ldots,O_m\}$. Each observation $O_m=\tuple{\boldsymbol{\gamma},\tau_i,t_{\tau_i},A_{\tau_i},R_{\tau_i},\xi_{\boldsymbol{\tau}}}$ is a six-tuple consisting of the following: a set of feature vectors $\boldsymbol{\gamma} = \{\gamma_{\tau_1},\gamma_{\tau_2},\ldots,\gamma_{\tau_n} \}$, where vector $\gamma_{\tau_j}$ describes the state of each task $\tau_j$; $\tau_i$, the task to be scheduled by the expert demonstrator at the current time step $t_{\tau_i}$; $A_{\tau_i} \subseteq \boldsymbol{A}$, the subset of agents allocated to task $\tau_i$ from the set of all agents $\boldsymbol{A}$; $R_{\tau_i} \subseteq \boldsymbol{R}$, the subset of resources allocated to task $\tau_i$ from the set of all resources $\boldsymbol{R}$; and $\xi_{\boldsymbol{\tau}}$, a set of context-specific and ``task-independent'' features that affect expert decision-making. The state feature vector for each task $\gamma_{\tau_j}$ incorporates features that affect the selection of the task for execution and may represent, for example, the deadline, the earliest time at which the task is available, the duration of the task, which resource $r$ the task requires, etc. The task-independent feature vector, $\xi_{\boldsymbol{\tau}}$, represents global state features, such as the proportion of agents that are currently idle. 

An \emph{agent} is defined as an entity that processes tasks and possesses the following set of attributes: time-varying physical location, travel speed, and task-specific proficiency (i.e., two agents may require different amounts of time to execute the same task). \color{black} A \emph{resource} is defined as an object required to process a task and possesses the following attributes: time-invariant physical location, a finite number of agents that can utilize the resource at any one time, and a task-specific proficiency (i.e., one resource may allow a task to be completed at a faster rate than another). In the event that no task is scheduled at time $t$, elements $\tau_i$, $A_{\tau_i}$, and $R_{\tau_i}$ in $O_m$ are null.

%


The goal is to learn a scheduling policy that selects a task ${\tau_i}$ to schedule at a selected time $t_{\tau_i}$ to be processed by agent a $a_{\tau_i}$ as a function of the task and problem state encoded by  $\gamma_{\tau_i}$ and $\xi_{\boldsymbol{\tau}}$. Our formulation assumes at least one agent is required to process one task, with the assignment and scheduling of agents to tasks determined by the scheduler. The assignment of a resource to a task is assumed to be either pre-allocated based on the problem specification or assigned by the scheduler. \color{black}

We assume that the cross product of the task-independent feature vectors and the task-dependent feature vector ($\xi_{\boldsymbol{\tau}} \times \gamma_{\tau_1} \times \gamma_{\tau_2} \times \ldots \times \gamma_{\tau_n}$) encodes sufficient information to make high quality scheduling decisions. Modeling choices may affect the dimensionalities of these feature vectors. For example, in one formulation the state of task $\tau_i$ may include a list of upper- and lowerbound temporal constraints between task $\tau_i$ and all other tasks $\tau_j$; alternatively, depending on the problem, a lower-dimensional representation of the same relevant information may simply include the latest possible time (i.e., the deadline) by which each task must start to satisfy the problem temporal constraints.

We note that our approach relies upon the ability of domain experts to articulate an appropriate set of features for the given problem. We believe this to be a reasonable limitation. Results from prior work have indicated that domain experts are adept at describing the high-level, contextual, and task-specific features used in their decision making; however, it is more difficult for experts to describe how they reason about these features \cite{Cheng:2006,Raghavan:2006}. In future work, we aim to extend our approach to include feature learning rather than relying upon experts to enumerate the important features they reason about in order to construct schedules.

Our learning approach de-constructs the problem into two steps: 1) For each agent, determine the candidate next task to schedule; and 2) For each candidate task, determine whether to schedule said task. 


\subsubsection{Learning Task Priorities}

In order to learn to correctly assign the next task to schedule, we transform each observation $O_m$ into a new set of observations by performing pairwise comparisons between the scheduled task $\tau_i$ and the set of unscheduled tasks (Equations \ref{eq:pairwisePos}-\ref{eq:pairwiseNeg}). Equation \ref{eq:pairwisePos} creates a positive example for each observation in which a task $\tau_i$ was scheduled. This example consists of the input feature vector, $\phi_{\tuple{\tau_i,\tau_x}}^m$, and a positive label, $y_{\tuple{\tau_i,\tau_x}}^m=1$. Each element of input feature vector $\phi_{\tuple{\tau_i,\tau_x}}^m$ is computed as the difference between the corresponding values in the feature vectors $\gamma_{\tau_i}$ and $\gamma_{\tau_x}$, describing scheduled task $\tau_i$ and unscheduled task $\tau_x$ concatenated with the high-level contextual feature vector $\xi_{\boldsymbol{\tau}}$. Equation \ref{eq:pairwiseNeg} creates a set of negative examples with $y_{\tuple{\tau_x,\tau_i}}^m=0$. For the input vector, we take the difference of the feature values between unscheduled task $\tau_x$ and scheduled task $\tau_i$ concatenated with the high-level contextual feature vector $\xi_{\boldsymbol{\tau}}$. 

We note that it is necessary to separate the task-independent features as point-wise terms so as to preserve their information. Consider the example task-independent feature, $\xi_{\boldsymbol{\tau}}^k$, representing the proportion of agents currently idle. If this feature would be encoded in each task-specific feature vector as $\gamma_{\tau_i}^k$, the result would be  $\gamma_{\tau_i}^k-\gamma_{\tau_j}^k=0$ for all tasks $\tau_i$ and $\tau_k$. Thus, for their information to be preserved for the learning algorithm, one must concatenate a separate vector of contextual features to the pairwise differences.


\begin{gather}
	^{rank}\theta^m_{\tuple{\tau_i,\tau_j}} := \left[ \xi_{\boldsymbol{\tau}},\gamma_{\tau_i}-\gamma_{\tau_j}\right],
	y^m_{\tuple{\tau_i,\tau_j}} = 1,
	\forall \tau_j \in \boldsymbol{\tau} \backslash \tau_i, \forall O_m \in \boldsymbol{O} | \tau_i \text{ scheduled in } O_m \label{eq:pairwisePos}\\ \nonumber  \\
	^{rank}\theta^m_{\tuple{\tau_j,\tau_i}} := \left[ \xi_{\boldsymbol{\tau}},\gamma_{\tau_j}-\gamma_{\tau_i}\right], 
	y^m_{\tuple{\tau_j,\tau_i}} = 0,
	\forall \tau_j \in \boldsymbol{\tau} \backslash \tau_i, \forall O_m \in \boldsymbol{O} | \tau_i \text{ scheduled in } O_m \label{eq:pairwiseNeg}
\end{gather}
\begin{gather}
	\widehat{\tau_i^*} = \argmax\limits_{\tau_i \in \boldsymbol{\tau}} \sum\limits_{\tau_j \in \boldsymbol{\tau}} f_{priority}\left(\tau_i,\tau_j\right)
	\label{eq:priorityFn}
\end{gather}
\normalsize

\begin{figure}
	\includegraphics[width =  0.975\textwidth]{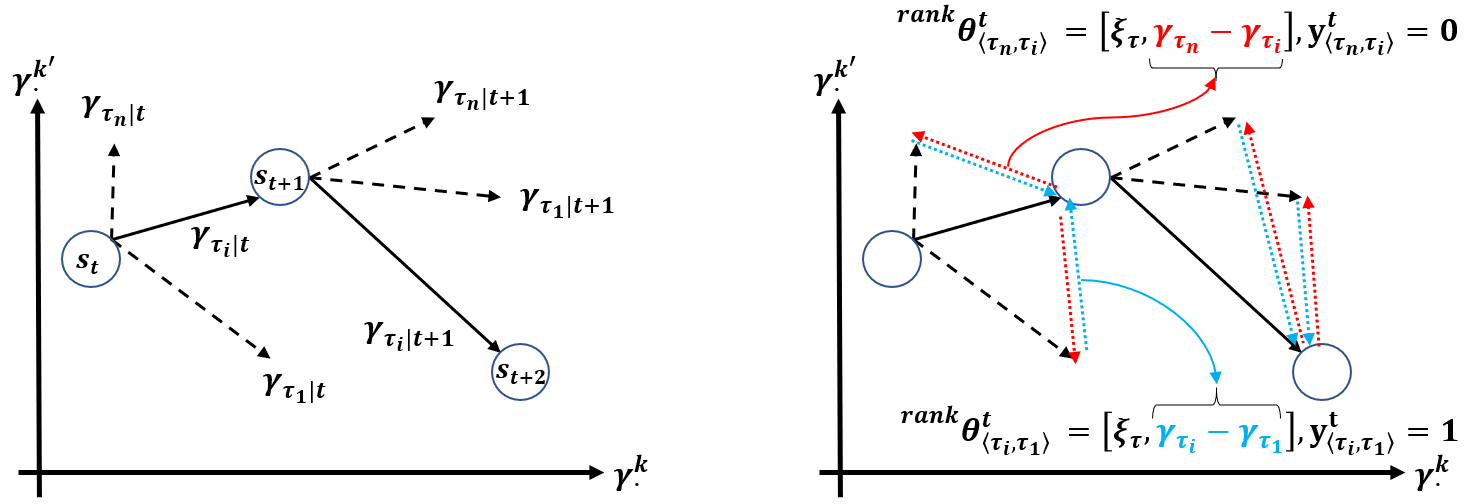}
	\caption{\color{black}This figure depicts the process for automatically generating positive and negative training examples for each $O_m \in \boldsymbol{O}$. The left graphic shows the apprentice scheduler's observations of the expert, and the right graphic depicts the construction of training examples through pairwise comparisons.\color{black}}
	\label{fig:FeatureSpace}
\end{figure}

\color{black}
Figure \ref{fig:FeatureSpace} is a graphical depiction of the process for automatically generating positive and negative training examples for each $O_m \in \boldsymbol{O}$. For illustrative purposes, the graphic depicts the process considering two task-specific features, $\gamma_{\cdot}^k$ and $\gamma_{\cdot}^{k'}$, corresponding to the x- and y-axes, respectively. 

In the left graphic, the node ``$s_t$" represents the state of the scheduling domain at time $t$, mapped to the feature space ($\gamma_{\cdot}^k$, $\gamma_{\cdot}^{k'}$). At this time $t$, the apprentice scheduler observes the demonstrator scheduling task $\tau_i$ (denoted by the solid arrow vector $\gamma_{\tau_i | t}$). The apprentice scheduler observes that the demonstrator chose to not schedule the two other available tasks $\tau_1$ or $\tau_n$ at $t$ (denoted by the dashed vectors $\gamma_{\tau_1 | t}$ and $\gamma_{\tau_n | t}$, respectively). After the scheduling and execution of $\tau_i$, the scheduling domain is observed to be in the state represented by node ``$s_{t+1}$". The figure shows the process repeating at time $t=1$. 

The right graphic depicts the generation of training examples. For each time step, the apprenticeship scheduler constructs positive and negative training examples through vector subtraction of task-dependent feature vectors. The red dotted lines depict the vector difference of the scheduled task's feature vector and each unscheduled task's feature vector; the resulting vectors are applied to construct negative training examples. The blue dotted lines depict the \emph{negative} vector difference of the scheduled task's feature vector and each unscheduled task's feature vector; the resulting vectors are applied to construct positive training examples. Recall that a contextual ``task-independent" feature vector, $\xi_{\boldsymbol{\tau}}$, is appended to each pairwise term in the formation of each training example $^{rank}\theta^m_{\tuple{\cdot,\cdot}}$. This procedure then repeats for each observation (i.e., each time step for each demonstrated schedule) and task. The value of this approach is that the learner does not need to explicitly solicit pairwise comparisons from the demonstrator; instead, the pairwise comparisons are derived automatically through observation of the expert demonstrator. \color{black}


\subsubsection{Learning to Schedule or Idle}

Given these observations $O_m$ and their associated features, we can train a classifier, $f_{priority}(\tau_i,\tau_x )\in\{0,1\}$, to predict whether it is better to schedule task $\tau_i$ as the next task rather than $\tau_x$. With this pairwise classifier, we can determine which single task $\tau_i$ is the highest-priority task $\tau_i^*$ according to Equation \ref{eq:priorityFn} by determining which task has the highest cumulative priority in comparison to the other tasks in $\boldsymbol{\tau}$. In this work, we train a single classifier, $f_{priority}(\tau_i,\tau_j )$, to model the behavior of the set of all agents rather than train one $f_{priority}(\tau_i,\tau_j)$ for each agent. $f_{priority}(\tau_i,\tau_j)$ is a function of all features associated with the agents; as such, agents need not be interchangeable, and different sets of features may be associated with each agent.   

Next, we must learn to predict whether $\tau_i^*$ should be scheduled or the agent should remain idle. To do so, we train a second classifier, $f_{act}(\tau_i )\in\{0,1\}$, that predicts whether or not $\tau_i$ should be scheduled. The observations set, $O$, consists either of examples in which a task was scheduled or those in which no task was scheduled. To train this classifier, we construct a new set of examples according to Equation \ref{eq:actFnc}, which assigns positive labels to examples from $O_m$ in which a task was scheduled and negative labels to examples in which no task was scheduled.
\begin{gather}
	^{act}\phi^m_{\tau_i} := \left[\xi_{\boldsymbol{\tau}},\gamma_{\tau_i}\right], 
	y_{\tau_i}^m = \left\{
	\begin{array}{lr}
		1 : \tau_i \text{ scheduled in } O_m \text{ } \land \tau_i \text{ scheduled in } O_{m+1} \\
		0 :  \tau_{\emptyset} \text{ scheduled in } O_m
	\end{array}
	\right.
	\label{eq:actFnc}
\end{gather} 

Finally, we construct a scheduling algorithm to act as an apprentice scheduler (Algorithm \ref{alg:apprenticeScheduler}). This algorithm takes as input the set of tasks, $\boldsymbol{\tau}$; agents, $\boldsymbol{A}$; temporal constraints (i.e., upper- and lowerbound temporal constraints) relating tasks in the problem, $\boldsymbol{TC}$; and the set of task pairs that require the same resources and can therefore not be executed at the same time, $\boldsymbol{\tau_R}$. Lines 1- 2 iterate over each agent at each time step. (In the event that resource-to-task assignments are not predefined, the algorithm would also iterate over each resource $r \in \boldsymbol{R}$ that could be assigned.) In Line 3, the highest-priority task, $\tau_i^*$, is determined for a particular agent. In Lines 4-5, $\tau_i^*$ is scheduled \emph{if} $f_{act}(\tau_i^*)$ predicts that $\tau_i^*$ should be scheduled at the current time. 

\begin{algorithm}
	\textbf{ApprenticeScheduler}($\boldsymbol{\tau}$,$\boldsymbol{A}$,$\boldsymbol{TC}$,$\boldsymbol{\tau_R}$)
	\begin{algorithmic}[1]
		\For{$t = 0$ to $T$}
		\For{all agents $a \in \boldsymbol{A}$}
		\State{$\tau_i^{*} \leftarrow \argmax\limits_{\tau_i \in \boldsymbol{\tau}} \sum\limits_{\tau_j \in \boldsymbol{\tau}} f_{priority}(\tau_i,\tau_j)$}
		\If{$f_{act}(\tau_i^*) == 1$}
		\State{Schedule $\tau_i^*$}
		\EndIf
		\EndFor
		\EndFor
	\end{algorithmic}
	\caption{Pseudocode for an Apprentice Scheduler}
	\label{alg:apprenticeScheduler}
\end{algorithm}

Note that iteration over agents (Line 2) can be performed according to a specific ordering, or the system can alternatively learn a more general priority function to select and schedule the best agent-task-resource tuple using $f_{priority}\left(\tuple{\tau_i,a,r},\tuple{\tau_j,a',r'}\right)$, $f_{act}\left(\tuple{\tau_i,a,r}^*\right)$. In the latter case, the features $\gamma_{\tau_i}$ are mapped to agent-task-resource tuples rather than tasks $\tau_i$, which represent the atomic (i.e., lowest-level) job. For the synthetic evaluation, we use the original formulation, $f_{priority}(\tau_i,\tau_j)$. For the ASMD application, we use $f_{priority}\left(\tuple{\tau_i^t,a,r},\tuple{\tau_j^t,a',r'}\right)$, where $\tau_i^t$ represents the objective of mitigating missile $i$ during time step $t$, $a$ is the decoy to be deployed, and $r$ is the physical location for that deployment. For the hospital domain evaluation, we use $f_{priority}\left(\tuple{\tau_i^j,a,r},\tuple{\tau_p^q,a',r'}\right)$, where $\tau_i^j$ represents the $j^{th}$ stage of labor for patient $i$, $a$ is the assigned nurse, and $r$ is the room to which the patient is assigned. For convenience in notation, we refer to this tuple as a ``scheduling action." \color{black} Finally, note that multiple agent-resource pairs can be assigned to a single task, $\tau_i$. The apprentice scheduler would first pick the best agent (or agent-resource pair) to assign to a task according to the $f_{priority}$ metric. During the same time step (or a subsequent time step), another agent (or agent-resource pair) can be added. The algorithm will continue to add assignments to the task until the null assignment (i.e., no further changes to the current set of assignments) is the best option according to $f_{act}$. \color{black}



Our model is a hybrid point- and pairwise formulation, which has several key benefits for learning to schedule form expert demonstration. First, we can directly apply standard classification techniques, such as a decision tree, support vector machine, logistic regression, or neural networks. Second, because this technique only considers two scheduling actions at a time, the model is non-parametric in the number of possible actions. Thus, the system can train on $f_{priority}(\tau_i,\tau_j)$ schedules with $a$ agents and $n$ tasks, yet apply $f_{priority}(\tau_i,\tau_j)$ to construct a schedule for a problem with $a'$ agents and $n'$ tasks where $a \neq a'$, $n \neq n'$, and $a*n \neq a'*n'$. Furthermore, it can even train $f_{priority}(\tau_i,\tau_j)$ on demonstrations of a heterogeneous data set of scheduling observations with differing numbers of agents and tasks. Third, the pairwise portion of the formulation provides structure for the learning problem. A formulation that simply concatenated the features of two or more scheduling actions would need to solve the more complex problem of learning the relationships between features and then how to use those relationships to predict the highest-priority scheduling action. Such a concatenation approach would suffer from the curse of dimensionality and require a very large training data set \cite{Indyk:1998}. Note, however, that this method requires the designer to appropriately partition the features into pairwise and pointwise components such that the pairwise portion does not lose information by considering the differences between actions' features. Fourth, the transformation of the observations into a pairwise model results in some features that are advantageous for learning from small data sets: the number of positive and negative training examples is balanced given that the algorithm simultaneously creates one negative label for every positive label, and the observations are bootstrapped to create $2*|\boldsymbol{\tau}|$ examples for each time step, rather than only $|\boldsymbol{\tau}|$ for a pointwise model, where $n = |\boldsymbol{\tau}|$. 

\section{Data Sets}
\label{sec:data}
Here, we validate that schedules produced by the learned policies are of comparable quality to those generated by human or synthetic experts. To do so, we considered a synthetic data set from the \textbf{XD [ST-SA-TA]} class of problems and two real-world data sets from the \textbf{CD [MT-MA-TA]} class of problems, as defined by Korsah et al. \cite{Korsah:2013}. We present each problem domain and describe the manner in which the data set of expert demonstrations for the domain was acquired. 


\subsection{Synthetic Data Set}
\label{sec:DataSet_VRP}
For our first investigation, we generated a synthetic data set of scheduling problems in which agents were assigned a set of tasks. The tasks were related through precedence or wait constraints, as well as deadline constraints, which could be absolute (relative to the start of the schedule) or relative to another task's initiation or completion time. Agents were required to access a set of shared resources to execute each task. Agents and tasks had defined starting locations, and task locations were static. Agents were only able to perform tasks when present at the corresponding task location, and each agent traveled at a constant speed between task locations. Task completion times were potentially non-uniform and agent-specific, as would be the case for heterogeneous agents. An agent that was incapable of performing a given task was assumed to have an infinite completion time for that task.  The objective was to minimize the makespan or other time-based performance measures.

This problem definition spans a range of scheduling problems, including the traveling salesman, job-shop scheduling, multi-vehicle routing and multi-robot task allocation problems, among others. We describe this range as a vehicle routing problem with time windows, temporal dependencies, and resource constraints (VRPTW-TDR), which falls within the \textbf{XD [ST-SA-TA]} class in the taxonomy by \citeA{Korsah:2013}: agents perform tasks sequentially (ST), each task requires one agent (SA), and commitments are made over time (TA). 

To generate our synthetic data set, we developed a mock scheduling expert that applies one of a set of context-dependent rules based on the composition of the given scheduling problem. This behavior was based upon rules presented in prior work addressing these types of problems \cite{GombolayRSS:2013,GombolayJAIS:2014,Solomon:1987,Tan:2001}. Our objective was to show that our apprenticeship scheduling algorithm learns both context-dependent rules and how to identify the associated context for their correct application. 

The mock scheduling expert functions as follows: First, the algorithm collects all alive and enabled tasks $\tau_i\in\boldsymbol{\tau_{AE}}$ as defined by \cite{Muscettola:1998}. Consider a pair of tasks, $\tau_i$ and $\tau_j$, with start and finish times $s_i,f_i$ and $s_j,f_j$, respectively, such that there is a wait constraint requiring $\tau_i$ to start at least $W_{\tuple{\tau_j,\tau_i}}$ units of time after $\tau_j$. A task $\tau_i$ is alive and enabled if $t \geq f_j + W_{\tau_j,\tau_i}$ for all such $\tau_j$ and $W_{\tuple{\tau_j,\tau_i}}$ in $\boldsymbol{\tau}$.

After task collection, the heuristic iterates over each agent to identify the highest-priority task, $\tau_i^*$, to schedule for that agent. The algorithm determines which scheduling rule is most appropriate to apply for each agent. If agent speed is sufficiently slow ($\leq 1$ m/s), travel time will become the major bottleneck. If agents move quickly but utilize one or more resources $R$ heavily ( $\sum_{\tau_i}\sum_{\tau_j} 1_{R_{\tau_i} = R_{\tau_j}} \geq \text{c}$ for some constant c), use of these resources can become the bottleneck. Otherwise, task durations and associated wait constraints are generally most important. 

If the algorithm identifies travel distance as the primary bottleneck, it chooses the next task by applying a priority rule well-suited for vehicle routing that minimizes a weighted, linear combination of features \cite{Gambardella:1999,Solomon:1987} comprised of the distance and angle relative to the origin between agent $a$ and $\tau_j$. This rule is depicted in Equation \ref{eq:VRPRule}, where $\vec{l}_x$ is the location of $\tau_j$, $\vec{l}_a$ is the location of agent $a$, $\theta_{xa}$ is the relative angle between the vector from origin to the agent location and the origin to the location of $\tau_j$, and $\alpha_1$ and $\alpha_2$ are weighting constants:
\begin{equation}
\tau_i^* \leftarrow \argmin\limits_{\tau_j \in \boldsymbol{\tau_{AE}}} \left( \|\vec{l}_x - \vec{l}_a\|+ \alpha_1 \theta_{xa} + \alpha_2 \|\vec{l}_x - \vec{l}_a\| \theta_{xa}\right)
\label{eq:VRPRule}
\end{equation}
If the algorithm identifies resource contention as the most important bottleneck, it employs a rule to mitigate resource contention in multi-robot, multi-resource problems based on prior work in scheduling for multi-robot teams \cite{GombolayRSS:2013}. Specifically, the algorithm uses Equation \ref{eq:RCRule} to select the high-priority task to schedule next, where $d_{\tau_j}$ is the deadline of $\tau_j$ and $\alpha_3$ is a weighting constant:
\begin{equation}
\tau_i^* \leftarrow \argmax\limits_{\tau_j \in \boldsymbol{\tau_{AE}}} \left(\left(\sum_{\tau_i}\sum_{\tau_j} 1_{R_{\tau_i} = R_{\tau_j}}\right) - \alpha_3 d_{\tau_j}\right)
\label{eq:RCRule}
\end{equation}
If the algorithm decides that temporal requirements are the major bottleneck, it employs an Earliest Deadline First rule (Equation \ref{eq:EDFRule}), which performs well across many scheduling domains \cite{Chen:2009,GombolayRSS:2013,GombolayJAIS:2014}:
\begin{equation}
\tau_i^* \leftarrow \argmin\limits_{\tau_j \in \boldsymbol{\tau_{AE}}} d_{\tau_j}
\label{eq:EDFRule}
\end{equation}
After selecting the most important task, $\tau_i^*$, the algorithm determines whether the resource required for $\tau_i^*$, $R_{\tau_i^{*}}$, is idle and whether the agent is able to travel to the task location by time $t$. If these constraints are satisfied, the heuristic schedules task $\tau_i^{*}$ at time $t$. (An agent is able to reach task $\tau_i^*$ if $t \geq f_j + k\left(l_i - l_j\right)/\|l_i - l_j\|$ for all $\tau_j \in \boldsymbol{\tau}$ that the agent has already completed, where $k$ is the agent's speed.)

We constructed the synthetic data set for two homogeneous agents and 20 partially ordered tasks located within a 20 x 20 grid. 

\subsection{Real-World Data Set: Anti-Ship Missile Defense}
\label{sec:ASMDDataSet}

In ASMD, the goal is to protect one's naval vessel against attacks by anti-ship missiles using ``soft-kill weapons" (i.e., decoys) that mimic the qualities of a target in order to direct the missile away from its intended destination. 


Developing tactics for soft-kill weapon coordination is highly difficult due to the relationship between missile behavior and soft-kill weapon characteristics. The control laws governing anti-ship missiles vary, and the captain must select the correct decoy types in order to counteract the associated anti-ship missiles. For example, a ship's captain may deploy a decoy that emits a large amount of heat in order to cause an enemy heat-seeking missile to fly toward the decoy rather than the ship. Also, an enemy missile may consider the spatial layout of all targets in order to select the nearest or furthest targets; in doing so, the missile may consider the magnitude of the radar reflectivity, radar emissions, and heat emissions, either separately or in various combinations. 

Further, decoys have different financial costs and timing characteristics: Some decoys, such as unmanned aerial vehicles (UAVs), are able to function throughout the entirety of an engagement, while others, such as an infrared (IR) flares, disappear after a certain time. As a result, a captain may be required to use multiple decoys in tandem in order to divert a single anti-ship missile, but may also be able to use a single decoy to defeat multiple missiles. There is a complex interplay between the types and locations of decoys relative to the control laws governing anti-ship missiles. For example, deployment of a particular decoy, while effective against one airborne enemy missile, may actually cause a second enemy missile that was previously homing in on a second decoy to now impact the ship.

The ASMD problem is characterized as the most complex class of scheduling problem according to the \citeA{Korsah:2013}. taxonomy : \textbf{CD [MT-MA-TA]}. The problem considers multi-task agents (MA) in the form of decoys, each of which can work to divert multiple missiles at the same time. The problem also incorporates multi-agent tasks (MT); a feasible solution may require the simultaneous use of multiple agents in order to complete an individual task. Further, time-extended agent allocation (TA) must be considered, given the potential future consequences of scheduling actions taken at the current moment. Finally, the ASMD problem falls within the CD class, because each task can be decomposed in a variety of ways -- each with their own cost -- in order to accomplish the same goal, with each decomposition affecting the value and feasibility of the decompositions of other tasks. \color{black} The full specification of the mixed-integer linear program formulation for the ASMD problem is provided in Appendix \ref{sec:AppendixASMD}.\color{black}



\begin{figure}
	\centering
	\includegraphics[width = 0.67\textwidth]{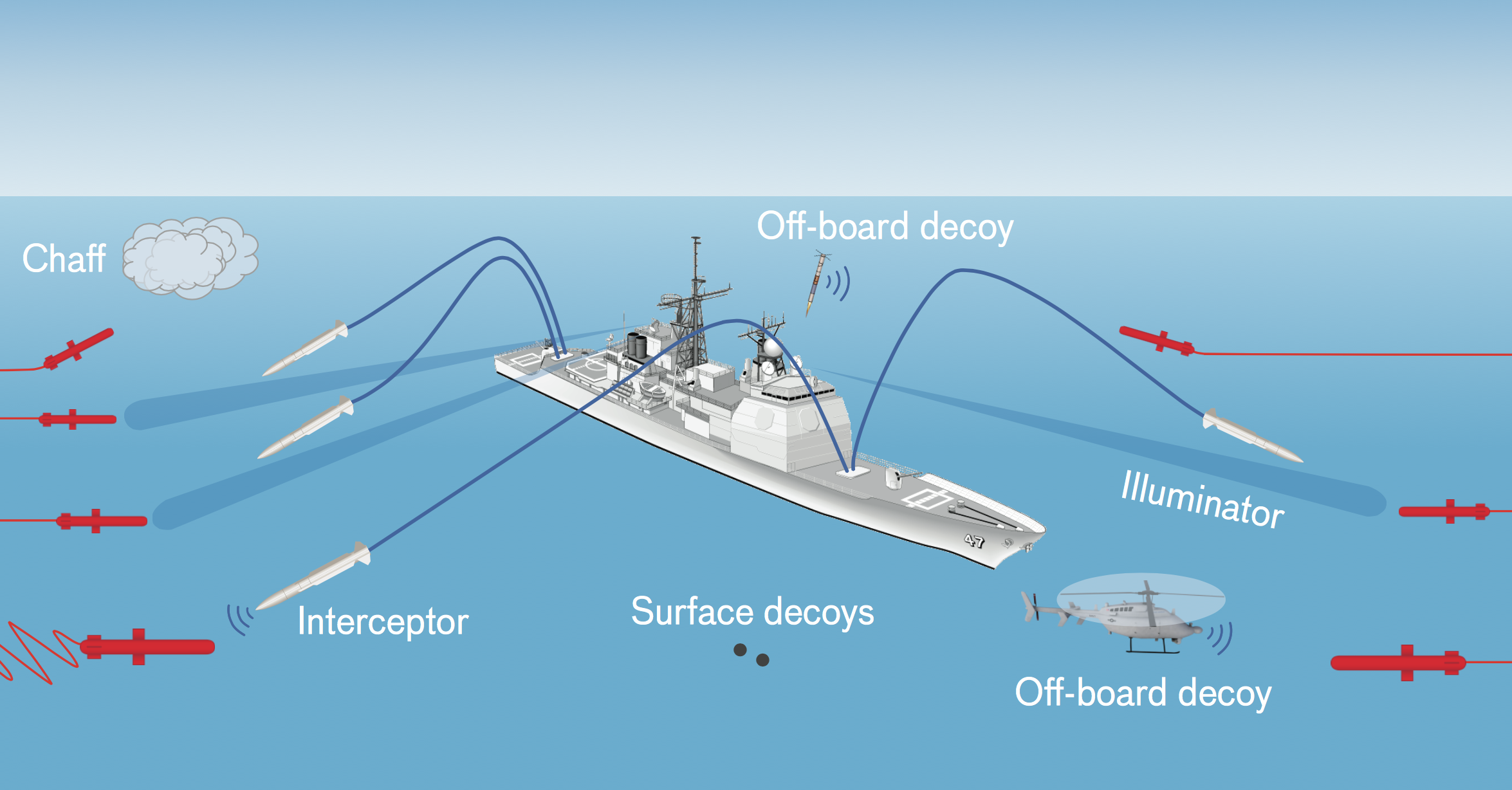}
	\caption{The tactical action officer aboard a naval vessel must coordinate a heterogeneous set of soft- and hard-kill weapons to defeat various anti-ship missiles.}
	\label{fig:SGDGraphic}
\end{figure}

\subsubsection{Data Collection}
A real-world data set was collected, consisting of human demonstrators of various skill levels solving the anti-ship missile defense (ASMD) weapon-to-target assignment problem.  Data was collected from domain experts playing a serious game, called Strike Group Defender\footnote{SGD was developed by Pipeworks Studio in Eugene, Oregon, USA.} (SGD), for ASMD training. Game scenarios involved five types of decoys and 10 types of threats. Threats were randomly generated for each played scenario, promoting the development of strategies that were robust to a varied distribution of scenarios. Each decoy had a specified effectiveness against each threat type. 

Players attempted to deploy a set of decoys by using the correct types at the correct locations and times in order to distract incoming missiles. Threats were launched over time; an effective deployment at time $t$ could become counterproductive in the future as new enemy missiles were launched. 

Games were scored as follows: $10,000$ points were received each time a threat was neutralized and $2$ points were received for each second a threat spent homing in on a decoy. Players lost $5,000$ points for each threat impact and $1$ point was deducted for each second a threat spent homing in on the player's ship. At each decoy deployment, players lost $25$-$1,000$ points depending upon decoy type.

The collected data set consisted of $311$ games played by $35$ humans across $45$ threat configurations, or ``levels." From this set, we also separately analyzed 16 threat configurations such that each configuration included at least one human demonstration in which the ship was successfully protected from all enemy missiles. For these 16 configurations, there were $162$ total games played by $27$ unique human demonstrators. The player cohort consisted of technical fellows and associates, as well as contractors at a federally funded research and development center (FFDRC), with expertise varying from ``generally knowledgeable about the ASMD problem" to ``domain experts" with professional experience or training in ASMD. 

\subsection{Real-World Data Set: Labor and Delivery}
\label{sec:LaborAndDeliveryDataSet}
\begin{figure}
	\centering
	\includegraphics[width = 1.0\textwidth]{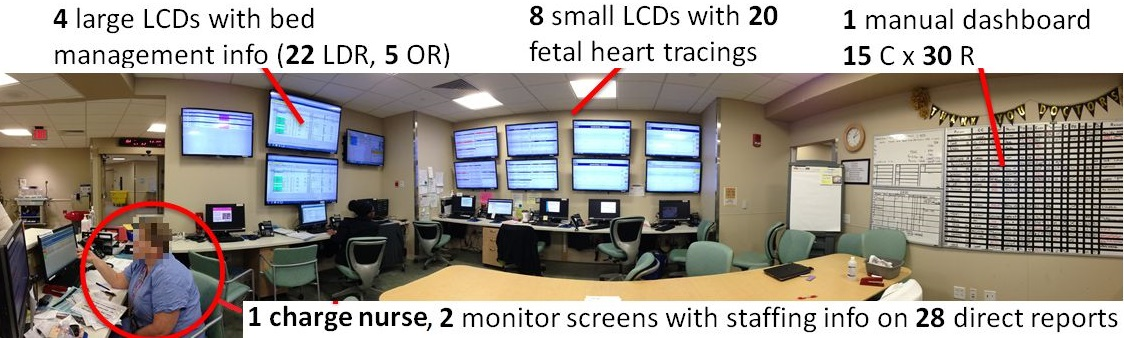}
	\caption{A resource nurse must assimilate a large variety and volume of information to effectively reason about resource management for patient care.}
	\label{fig:resourceNurseDiagram1}
\end{figure} 

To further evaluate our approach, we applied our method to a second data set collected from a labor and delivery floor at a Boston hospital. In this domain, a ``resource nurse'' must solve a problem of task allocation and schedule optimization with stochasticity in the number and types of patients and the duration of tasks. Specifically, the resource nurse is responsible for ensuring that the correct patient is in the correct type of room at the correct time, with the correct types of nurses present to care for those patients. The functions of a resource nurse are to assign nurses to take care of labor patients; assign patients to labor beds, recovery room beds, operating rooms, antepartum ward beds or postpartum ward beds; assign scrub technicians to assist with surgeries in operating rooms; call in additional nurses if necessary; accelerate, delay or cancel scheduled inductions or cesarean sections; expedite active management of a patient in labor; and reassign roles among nurses. 

Using our apprenticeship scheduling method in for the Labor and Delivery problem domain, a task $\tau_i$ represents the set of steps (subtasks) required to care for patient $i$, and each $\tau_i^j$ is a given stage of labor for that patient. Stages of labor are related by stochastic lowerbound constraints $W_{\tuple{\tau_i^j,\tau_x^y}}$, requiring the stages to progress sequentially. There are stochastic time constraints, $D^{abs}_{\tau_i^j}$ and $D^{rel}_{\tuple{\tau_i^j,\tau_x^y}}$, relating the stages of labor to account for the inability of resource nurses to perfectly control when a patient will move from one stage to the next. Arrivals of $\tau_i$ (i.e. patients) are drawn from stochastic distributions. The model considers three types of patients: scheduled cesarean patients, scheduled induction patients and unscheduled patients.  The set of $W_{\tuple{\tau_i^j,\tau_x^y}}$, $D^{abs}_{\tau_i^j}$ and $D^{rel}_{\tuple{\tau_i,\tau_j}}$ are dependent upon patient type.

Labor nurses are modeled as agents with a finite capacity to process tasks in parallel, where each subtask requires a variable amount of this capacity. For example, a labor nurse may generally care for a maximum of two patients simultaneously. If the nurse is caring for a patient who is ``full and pushing" (i.e., the cervix is fully dilated and the patient is actively trying to push out the baby) or in the operating room, he or she may only care for that patient. 

Rooms on the labor floor (e.g., a labor room, an operating room, etc.) are modeled as resources, which process subtasks in series. Agent and resource assignments to subtasks are pre-emptable, meaning that the agent and resource assigned to care for any patient during any step in the care process may be changed over the course of executing that subtask.

In this formulation, $\tensor*[^t]{A}{^a_{\tau_i^j}}\in\{0,1\}$ is a binary decision variable for assigning agent $a$ to subtask $\tau_i^j$ for time epoch $[t,t+1)$. $\tensor*[^t]{G}{^a_{\tau_i^j}}$ is an integer decision variable for assigning a certain portion of the effort of agent $a$ to subtask $\tau_i^j$ for time epoch $[t,t+1)$. $\tensor*[^t]{R}{^r_{\tau_i^j}}\in\{0,1\}$ is a binary decision variable for whether subtask $\tau_i^j$ is assigned resource $r$ for time epoch $[t,t+1)$. $H_{\tau_i} \in \{0,1\}$ is a binary decision variable for whether task $\tau_i$ and its corresponding subtasks are to be completed. $U_{\tau_i^j}$ specifies the effort required from any agent to work on $\tau_i^j$. $s_{\tau_i^j}, f_{\tau_i^j}\in [0,\infty)$ are the start and finish times of $\tau_i^j$.

\normalsize\begin{eqnarray}
&\min fn\left(\{\tensor*[^t]{A}{^a_{\tau_i^j}}\},\{\tensor*[^t]{G}{^a_{\tau_i^j}}\}, \{\tensor*[^t]{R}{^r_{\tau_i^j}}\}, \{H_{\tau_i}\}, \{s_{\tau_i^j},f_{\tau_i^j}\}\right)
\label{eq:objectiveNurseGeneral}
\end{eqnarray}
\begin{align}
	\sum_{a \in A} \tensor*[^t]{A}{^a_{\tau_i^j}} &\geq 1- M\left(1-H_{\tau_i}\right),\forall \tau_i^j \in \boldsymbol{\tau}, \forall t
	\label{eq:eachPatientGetsANurse} \\
	M  \left( 2-\tensor*[^t]{A}{^a_{\tau_i^j}}-H_{\tau_i}\right) &\geq 
	-U_{\tau_i^j} + \tensor*[^t]{G}{^a_{\tau_i^j}} \geq \nonumber \\ M\left(\tensor*[^t]{A}{^a_{\tau_i^j}}+H_{\tau_i}-2\right), \forall \tau_i^j \in \boldsymbol{\tau}, \forall t
	\label{eq:eachPatinetGetsEnoughNursing1} \\
	\sum_{\tau_i^j \in \boldsymbol{\tau}} \tensor*[^t]{G}{^a_{\tau_i^j}} &\leq C_a, \forall a \in A, \forall t
	\label{eq:agentCapacity} \\
	\sum_{r \in R} \tensor*[^t]{R}{^r_{\tau_i^j}} &\geq 1 - M\left(1-H_{\tau_i}\right), \forall \tau_i^j \in \boldsymbol{\tau}, \forall t
	\label{eq:fullResourceAssign}
\end{align}
\begin{align}
	\sum_{\tau_i^j \in \boldsymbol{\tau}} \tensor*[^t]{R}{^r_{\tau_i^j}} &\leq 1, \forall r \in R, \forall t
	\label{eq:resourceCapacity}\\
	ub_{\tau_i^j} \geq f_{\tau_i^j} - s_{\tau_i^j} &\geq lb_{\tau_i^j}, \forall \tau_i^j \in \boldsymbol{\tau}
	\label{eq:taskUBLBNurse}\\
	s_{\tau_x^y} - f_{\tau_i^j} &\geq W_{\tuple{\tau_i,\tau_j}}, \forall \tau_i, \tau_j \in \boldsymbol{\tau} |, \forall W_{\tuple{\tau_i,\tau_j}} \in \boldsymbol{TC}
	\label{eq:WaitConstraintNurse} \\
	f_{\tau_x^y} - s_{\tau_i^j} &\leq D^{rel}_{\tuple{\tau_i,\tau_j}}, \forall \tau_i, \tau_j \in \boldsymbol{\tau} | \exists D^{rel}_{\tuple{\tau_i,\tau_j}} \in \boldsymbol{TC}
	\label{eq:RelativeDeadlineNurse} \\
	f_{\tau_i^j} &\leq D^{abs}_{\tau_i}, \forall \tau_i \in \boldsymbol{\tau} | \exists D^{abs}_{\tau_i} \in \boldsymbol{TC}
	\label{eq:AbsoluteDeadlineNurse}
\end{align}
\normalsize
Equation \ref{eq:eachPatientGetsANurse} enforces that each subtask $\tau_i^j$ during each time epoch $[t,t+1)$ is assigned a single agent. Equation \ref{eq:eachPatinetGetsEnoughNursing1} ensures that each subtask $\tau_i^j$ receives a sufficient portion of the effort of its assigned agent $a$ during epoch $[t,t+1)$. Equation \ref{eq:agentCapacity} ensures that agent $a$ is not oversubscribed. 
Equation \ref{eq:fullResourceAssign} ensures that each subtask $\tau_i^j$ of each task $\tau_i$ that is to be completed (i.e., $H_{\tau_i}=1$) is assigned one resource $r$. Equation \ref{eq:resourceCapacity} ensures that each resource $r$ is assigned to only one subtask during each epoch $[t,t+1)$. Equation \ref{eq:taskUBLBNurse} requires the duration of subtask $\tau_i^j$ to be less than or equal to $ub_{\tau_i^j}$ and at least $lb_{\tau_i^j}$ units of time. Equation \ref{eq:WaitConstraintNurse} requires that $\tau_x^y$ occurs at least $W_{\tuple{\tau_i^j,\tau_x^y}}$ units of time after $\tau_i^j$. Equation \ref{eq:RelativeDeadlineNurse} requires that the duration between the start of $\tau_i^j$ and the finish of $\tau_x^y$ be less than $D^{rel}_{\tuple{\tau_i^j,\tau_x^y}}$. Equation \ref{eq:AbsoluteDeadlineNurse} requires that $\tau_i^j$ finishes before $D^{abs}_{\tau_i^j}$ units of time have expired since the start of the schedule. 

The functions of a resource nurse are to assign nurses to take care of labor patients and to assign patients to labor beds, recovery room beds, operating rooms, antepartum ward beds or postpartum ward beds. The resource nurse has substantial flexibility when assigning beds, and his or her decisions will depend upon the type of patient and the current status of the unit in question. He or she must also assign scrub technicians to assist with surgeries in operating rooms, and call in additional nurses if required. The corresponding decision variables for staff assignments and room/ward assignments in the above formulation are $\tensor*[^t]{A}{^a_{\tau_i^j}} $ and $\tensor*[^t]{R}{^r_{\tau_i^j}}$, respectively.  

The resource nurse may accelerate, delay or cancel scheduled inductions or cesarean sections in the event that the floor is too busy. Resource nurses may also request expedited active management of a patient in labor. The decision variables for the timing of transitions between the various steps in the care process are described by $s_{\tau_i^j}$ and $f_{\tau_i^j}$. The commitments to a patient (or that patient's procedures) are represented by $H_{\tau_i}$. 

The resource nurse may also reassign roles among nurses: For example, a resource nurse may pull a nurse from triage, or even care for patients herself if the floor is too busy. Or, if a patient's condition is particularly acute (e.g., the patient has severe preeclampsia), the resource nurse may assign one-to-one nursing. The level of attentional resources a patient requires and the level a nurse has available correspond to variables $U_{\tau_i^j} $ and $\tensor*[^t]{G}{^a_{\tau_i^j}}$, respectively. The resource nurse makes his or her decisions while considering current patient status $\Lambda_{\tau_i^j}$, which is manually transcribed on a whiteboard, as shown in Figure \ref{fig:resourceNurseDiagram1}.

The stochasticity of the problem arises from the uncertainty in the upper- and lowerbound of the durations $(ub_{\tau_i^j} \text{ and } lb_{\tau_i^j})$ of each of the steps in caring for a patient; the number and types of patients, $\boldsymbol{\tau}$; and the temporal constraints, $\boldsymbol{TC}$, relating the start and finish of each step. These variables are a function of the resource and staff allocation variables, $\tensor*[^t]{R}{^a_{\tau_i^j}} and \tensor*[^t]{A}{^a_{\tau_i^j}}$, as well as patient task state $\Lambda_{\tau_i^j}$, which includes information on patient type (i.e., presentation with scheduled induction, scheduled cesarean section, or acute unplanned anomaly), gestational age, gravida, parity, membrane status, anesthesia status, cervix status, time of last exam and the presence of any comorbidities. Formally, $\left(\{ub_{\tau_i^j}, lb_{\tau_i^j} | \tau_i^j \in \boldsymbol{\tau}\}, \boldsymbol{\tau},\boldsymbol{TC}\right) \sim P(\{\tensor*[^t]{R}{^a_{\tau_i^j}}, \tensor*[^t]{A}{^a_{\tau_i^j}}, \Lambda_{\tau_i^j}, \forall t \in [0,1,\ldots,T] \} )$. 

The computational complexity of completely searching for a solution that satisfies the constraints in Equations \ref{eq:eachPatientGetsANurse}-\ref{eq:AbsoluteDeadlineNurse} is given by $O\left(2^{|A||R|T^2}C_a^{|A|T}\right)$, where $|A|$ is the number of agents, with each agent possessing an integer processing capacity of $C_a$. There are $n$ tasks $\tau_i$, each with $m_i$ subtasks, $|R|$ resources, and an integer-valued planning horizon of $T$ units of time. In practice, there are $\sim$ $10$ nurses (agents) who can care for up to two patients at a time (i.e., $C_a = 2, \forall a \in A$), $20$ different rooms (resources) of varying types, $20$ patients (tasks) at any one time, and a planning horizon of $12$ hours or $720$ minutes, yielding a worst-case complexity of $\sim 2^{10*20*720^2}2^{10*720} \geq 2^{10^6}$, which is computationally intractable for exact methods without the assistance of informative search heuristics.


\subsubsection{Data Collection}
\label{sec:NurseDataSet}
\begin{sidewaysfigure*}
	\centering
	\includegraphics[scale = 0.4]{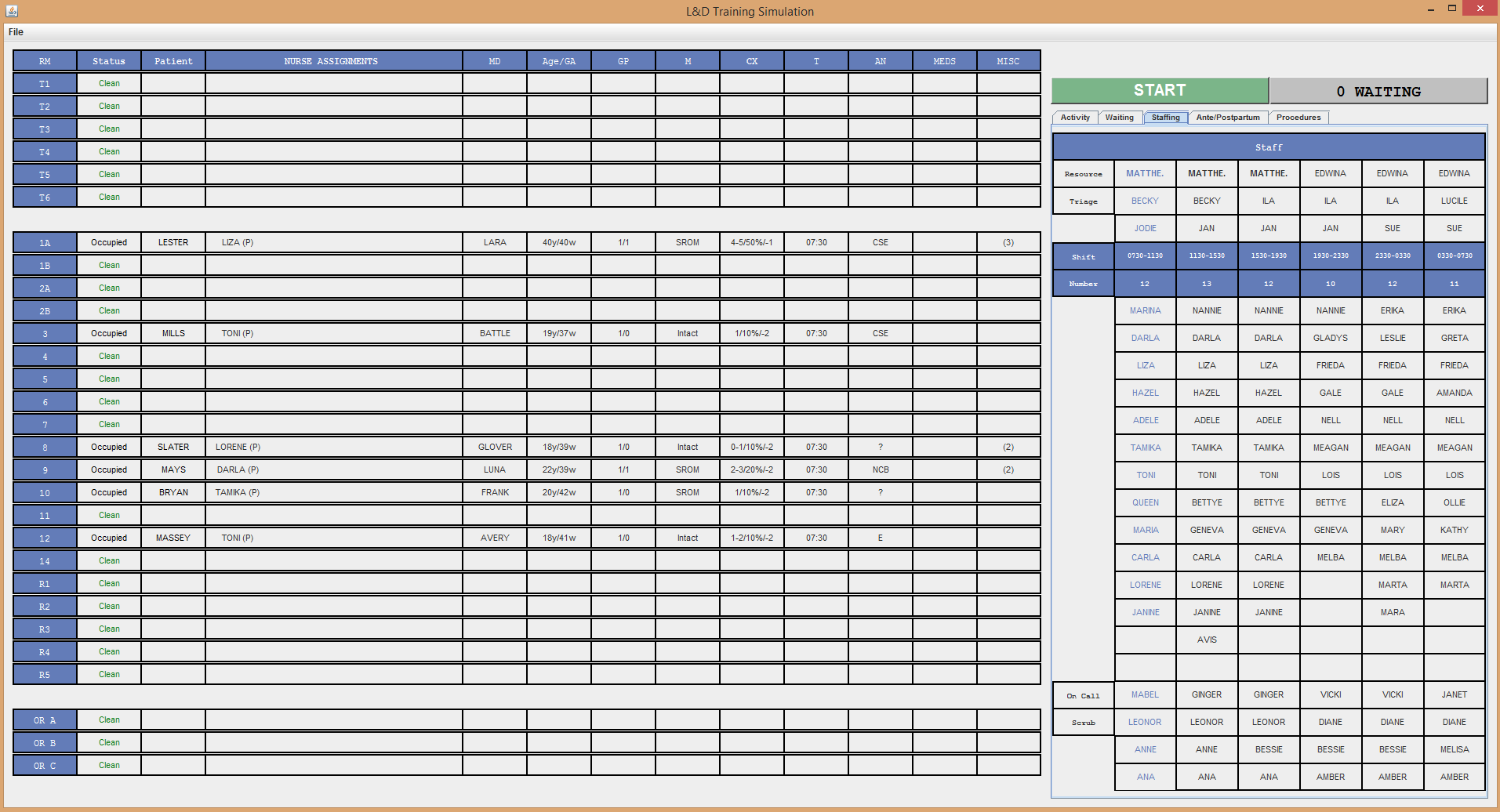}
	\caption{A screen capture of our simulation of a labor and delivery floor.}
	\label{fig:LDSimulation}
\end{sidewaysfigure*}

To collect data from resource nurses about their decisions, a high-fidelity simulation of a labor and delivery floor was developed, as depicted in Figure \ref{fig:LDSimulation}. We developed this simulation in collaboration with Beth Israel Medical Deaconess Hospital in Boston. The effort was part of a quality-improvement project at the hospital to develop training tools and involved a rigorous, year-long design and iteration process that included workshops with nurses, physicians, and medical students to ensure the tool accurately captured the role of a resource nurse. Parameters within the simulation (e.g., patient arrivals, timelines for labor progression) were drawn from medical textbooks and papers and modified through alpha and beta testing to ensure that the simulation closely mirrored the patient population and nurse experience at our partner hospital.

We invited expert resource nurses to play this simulation in order to collect a data set for training our apprenticeship scheduling algorithm. This data set was generated by seven resource nurses working with the simulation for a total of $2 \sfrac{1}{2}$ hours, simulating $60$ hours of elapsed time on a real labor floor and yielding a set of more than $3,013$ individual decisions.

\section{Empirical Evaluation of Apprenticeship Scheduling}
In this section, we evaluate our prototype for apprenticeship scheduling using synthetic and real-world data sets.

\subsection{Synthetic Data Set}
\label{sec:syntheticDataSet}
We trained our model using a decision tree, KNN classifier, logistic regression (logit) model, a support vector machine with a radial basis function kernel (SVM-RBF), and a neural network to learn $f_{priority}(.,.)$ and $f_{act}(.)$. We randomly sampled $85\%$ of the data for training and $15\%$ for testing. 

We defined the input features as follows: The high-level feature vector of the task set, $\xi_\tau$, was comprised of the agents' speed and the degree of resource contention, $\sum_{\tau_i}\sum_{\tau_j} 1_{R_{\tau_i} = R_{\tau_j}}$. The task-specific feature vector, $\gamma_{\tau_i}$, was comprised of the task's deadline, a binary indicator for whether or not the task's precedence constraints had been satisfied, the number of other tasks sharing the given task's resource, a binary indicator for whether or not the given task's resource was available, the travel time remaining to reach the task location, the distance agent $a$ would travel to reach $\tau_i$, and the angular difference between the vector describing the location of agent $a$ and the vector describing the position of $\tau_i$ relative to agent $a$.

We compared the performance of our pairwise approach with pointwise and na\"{i}ve approaches. In the pointwise approach, training examples for selecting the highest-priority task were of the form ${^{rank}}\phi_{\tau_i}^{m} := [\xi_{\boldsymbol{\tau}},\gamma_{\tau_i}]$. The label $\gamma_{\tau_i}^m$ was equal to $1$ if task $\tau_i$ was scheduled in observation $m$, and was $0$ otherwise. In the na\"{i}ve approach, examples were comprised of an input vector that concatenated the high-level features of the task set and the task-specific features of the form $^{rank}\phi^m := [\xi_{\boldsymbol{\tau}},\gamma_{\tau_1},\gamma_{\tau_2},\ldots,\gamma_{\tau_n}]$; labels $y^m$ were given by the index of the task $\tau_i$ scheduled in observation $m$.

\begin{figure}
	\centering
	\begin{subfigure}{.5\textwidth}
		\centering
		\includegraphics[width = 0.9\textwidth]{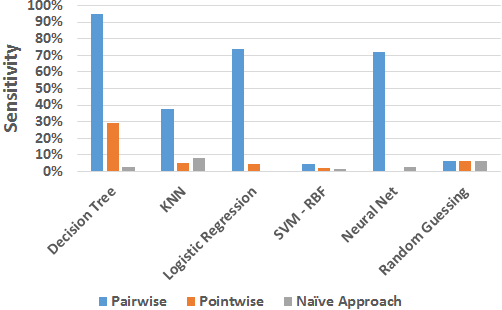}
		\caption{Sensitivity}
		\label{fig:sensitivity}
	\end{subfigure}%
	\begin{subfigure}{.5\textwidth}
		\centering
		\includegraphics[width = 0.9\textwidth]{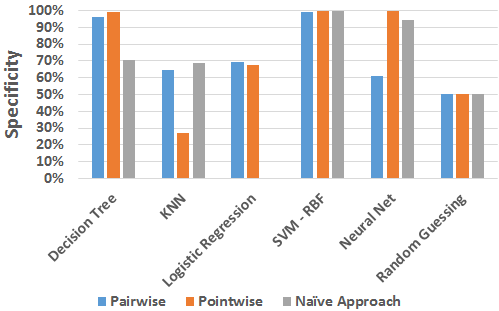}
		\caption{Specificity}
		\label{fig:specificity}
	\end{subfigure}
	\caption{Figures \ref{fig:sensitivity}-\ref{fig:specificity} depict the sensitivity and specificity of ML techniques using the pairwise, pointwise and na\"{i}ve approaches.}
	\label{fig:test1}
\end{figure}

Figures \ref{fig:sensitivity}-\ref{fig:specificity} depict the sensitivity (true positive rate) and specificity (true negative rate), respectively, of the model. We found that a pairwise model outperformed the pointwise and na\"{i}ve approaches. Within the pairwise model, a decision tree yielded the best performance: The trained decision tree was able to identify the correct task and when to schedule that task $95\%$ of the time, and was able to accurately predict when no task should be scheduled $96\%$ of the time. 

\begin{figure}
	\centering
	\begin{subfigure}{.5\textwidth}
		\centering
		\includegraphics[width=0.9\textwidth]{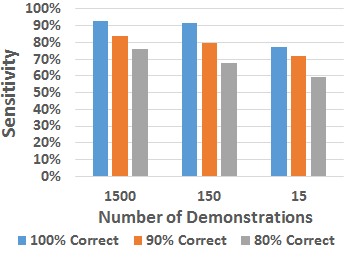}
		\caption{Sensitivity}
		\label{fig:sensitivity_noisy}
	\end{subfigure}%
	\begin{subfigure}{.5\textwidth}
		\centering
		\includegraphics[width=0.9\linewidth]{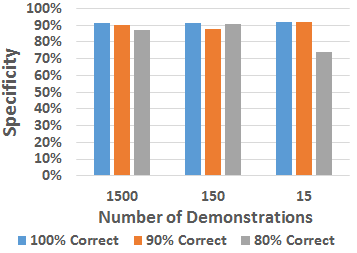}
		\caption{Specificity}
		\label{fig:specificity_noisy}
	\end{subfigure}
	\caption{Figures \ref{fig:sensitivity_noisy}-\ref{fig:specificity_noisy} depict the sensitivity and specificity of a pairwise decision tree, varying the number and proportion of correct demonstrations.}
	\label{fig:test2}
\end{figure}

To more fully understand the performance of a decision tree trained with a pairwise model as a function of the number and quality of training examples, we trained decision trees with the pairwise model using 15, 150, and 1,500 demonstrations. The sensitivity and specificity depicted in Figures \ref{fig:sensitivity_noisy} and \ref{fig:specificity_noisy} for 15 and 150 demonstrations represent the mean sensitivity and specificity of 10 models trained via random sub-sampling without replacement. 

We also varied the quality of the training examples, assuming the demonstrator was operating under an $\epsilon$-greedy approach with a $(1-\epsilon)$ probability of selecting the correct task to schedule, and selecting another task from a uniform distribution otherwise. \color{black} 
Our goal in this evaluation was to empirically investigate the impact of noisy demonstrations (i.e., those in which the demonstrator does not always select the``best'' tasks) on the quality of the learned policy. There are a number of possible models for introducing such noise, including an epsilon-greedy approach or a softmax model. An epsilon-greedy approach is expected to produce lower-quality demonstrations compared with a noisy human demonstrator, since a human would be more likely to select the second- or third-best task when making an error than to select a task at random, thus making the LfD problem more difficult. While no model will perfectly imitate an imperfect human demonstrator, we selected an epsilon-greedy approach as a reasonably conservative method of introducing more noise than might be generated by an imperfect human demonstrator. 
\color{black}

Training a model from pairwise comparisons of between the scheduled and each unscheduled tasks produced a comparable policy to that of the synthetic expert. The decision tree model performed well due to the modal nature of the multifaceted scheduling heuristic. Note that this data set consisted of scheduling strategies with mixed discrete-continuous functional components; performance could potentially be improved upon in future work by combining decision trees with logistic regression. This hybrid learning approach has been successful in prior ML classification tasks~\cite{Landwehr:2005} and can be readily applied to this apprenticeship scheduling framework. There is also an opportunity to improve performance through hyperparameter tuning (e.g., to select the minimum number of examples in each leaf of the decision tree). We leave comprehensive investigation of the relative benefits for a range of learning techniques for future work.

Note that the results presented in Figures \ref{fig:sensitivity}-\ref{fig:specificity_noisy} were achieved without any hyperparameter tuning. For example, with the decision tree, we did not perform an inner cross-validation loop to estimate the minimum number of examples in each leaf to achieve the best performance. The purpose of this analysis was to show that, with our pairwise approach, the system can accurately learn expert heuristics from example. In the following section, we investigate how apprenticeship scheduling using a decision tree classifier can be improved upon via an inner cross-validation loop to tune the model's hyperparameters.

\subsubsection{Performance of Decision Tree with Hyperparameter Tuning}

We performed our initial analysis, detailed above, to identify which techniques have inherent advantages that can be realized without extensive hyperparameter tuning. Our results indicate that the pairwise formulation for apprenticeship scheduling, in conjunction with a decision tree classifier, has advantages over alternative formulations for learning a high-quality scheduling policy. Given evidence of this advantage, we further evaluated the potential of the pairwise formulation with hyperparameter tuning.

To improve the performance of the model, we manipulated the ``leafiness" of the decision tree to find the best setting to increase the accuracy of the apprenticeship scheduler. Specifically, we varied the minimum number of training examples required in each leaf of the tree. As the minimum number required for each leaf decreases, the chance of over-fitting to the data increases. Conversely, as the minimum number increases, the chance of not learning a helpful policy (under-fitting) increases. To identify the best number of leaves for generalization, we tested values for the minimum number of examples required for each leaf of the decision tree in the set $\{1,5,10,25,50,100,250,500,1000\}$. If the minimum number of examples in each leaf exceeded the total number of examples, the setting was trivially set to the total number of examples available for training.

We performed $5$-fold cross-validation for each value of examples as follows: We trained an apprentice scheduler on four-fifths of the training data and tested on one-fifth of the data, and recorded the average testing accuracy across each of the five folds. Then, we used the setting of the minimum number of examples required for each leaf that yielded the best accuracy during cross-validation to train a full apprenticeship scheduling model on all of the training data ($85\%$ of the total data). Finally, we tested the full apprenticeship scheduling model on the $15\%$ of the total data reserved for testing. Thus, none of the data used to test the full model was used to estimate the best setting for the leafiness of the tree. We repeated this procedure 10 times, randomly sub-sampling the data and taking the average performance across the 10 trials. 

\begin{figure}
	\centering
	\begin{subfigure}{.5\textwidth}
		\centering
		\includegraphics[width=0.9\textwidth]{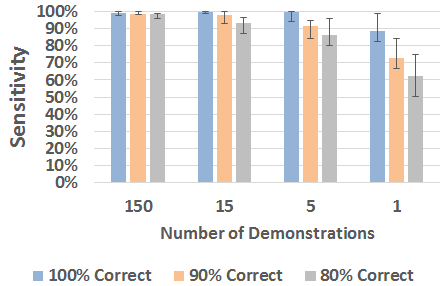}
		\caption{Sensitivity}
		\label{fig:sensitivity_noisy_new_homo}
	\end{subfigure}%
	\begin{subfigure}{.5\textwidth}
		\centering
		\includegraphics[width=0.9\linewidth]{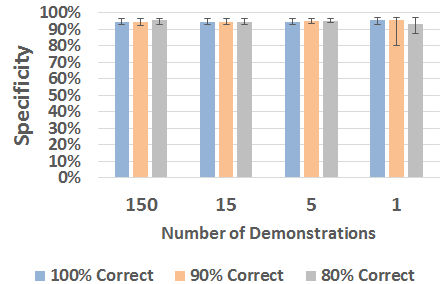}
		\caption{Specificity}
		\label{fig:specificity_noisy_new_homo}
	\end{subfigure}
	\caption{Figures \ref{fig:sensitivity_noisy_new_homo}-\ref{fig:specificity_noisy_new_homo} depict the sensitivity and specificity for a pairwise decision tree tuned for leafiness, varying the number and proportion of correct demonstrations. The corresponding data set comprised schedules with homogeneous agents.}
	\label{fig:test3}
\end{figure}

\begin{figure}
	\centering
	\begin{subfigure}{.475\textwidth}
		\centering
		\includegraphics[width=\linewidth]{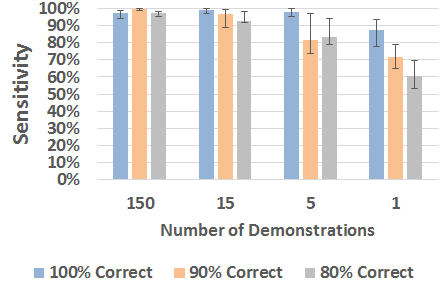}
		\caption{Sensitivity}
		\label{fig:sensitivity_noisy_new_hetero}
	\end{subfigure}%
	\begin{subfigure}{.475\textwidth}
		\centering
		\includegraphics[width=\linewidth]{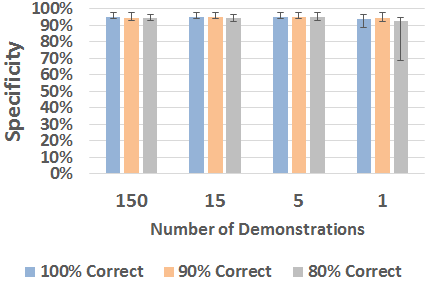}
		\caption{Specificity}
		\label{fig:specificity_noisy_new_hetero}
	\end{subfigure}
	\caption{Figures \ref{fig:sensitivity_noisy_new_hetero}-\ref{fig:specificity_noisy_new_hetero} depict the sensitivity and specificity for a pairwise decision tree tuned for leafiness, varying the number and proportion of correct demonstrations. The corresponding data set comprised schedules with heterogeneous agents.}
	\label{fig:test4}
\end{figure}

The sensitivity and specificity of the fully trained apprenticeship scheduling algorithm are depicted in Figures \ref{fig:sensitivity_noisy_new_homo} and \ref{fig:specificity_noisy_new_homo} for 1, 5, 15, and 150 scheduling demonstrations with homogeneous agents, and in Figures \ref{fig:sensitivity_noisy_new_hetero} and \ref{fig:specificity_noisy_new_hetero} for demonstrations with heterogeneous agents. As before, we also varied the quality of the training examples, assuming the demonstrator was operating under an $\epsilon$-greedy approach with a $(1-\epsilon)$ probability of selecting the correct task to schedule and selecting another task from a uniform distribution otherwise. 

For both the homogeneous and heterogeneous cases, we found that the apprenticeship scheduling algorithm was able to average $\geq 90\%$ sensitivity and specificity either with five perfect schedules or 15 schedules generated by an operator making mistakes $20\%$ of the time. Hyperparameter tuning substantially increased the sensitivity of the model from $59\%$ to $82\%$ for five scheduling examples generated by an operator making mistakes $20\%$ of the time. (Recall that a schedule consists of allocating 20 tasks to two workers and sequencing those tasks in time.)

Through our synthetic evaluation, we have shown that our apprentice scheduling algorithm is able to learn to make sequential decisions that accurately emulate the decision making process of a mock expert. The apprenticeship scheduler model shows a robust ability to learn from sparse, noisy data. 
In the following sections, we investigate the ability of the apprentice scheduler to learn from scheduling demonstrations produced by experts performing real-world scheduling tasks. 

\subsection{Real-World Data Set: ASMD}
\label{sec:ASMD_Results}
We trained a decision tree with our pairwise scheduling model and tested its performance via leave-one-out cross-validation involving 16 real demonstrations in which a player successfully protected the ship from all enemy missiles. Each demonstration originated from a unique threat scenario. Features for each decoy/missile pair (or null decoy deployment due to inaction) included indicators for whether a decoy had been placed such that a missile was successfully distracted by that decoy, whether a missile would be lured into hitting the ship due to decoy placement, or whether a missile would be unaffected by decoy placement. 

Across all 16 scenarios, the mean player score was $74,728$ $\pm$ $26,824$. With only 15 examples of expert human demonstrations, our apprenticeship scheduling model achieved a mean score of $87,540$, with a standard deviation of $16,842$. We hypothesized that scores produced by the learned policy would be statistically significantly better than the scores achieved by the human demonstrators. The null hypothesis stated that the number of scenarios in which the apprenticeship scheduling model achieved superior performance would be less than or equal to the number of scenarios in which the mean score of the human demonstrators was superior to that of the apprenticeship scheduler. We set the significance level at $\alpha = 0.05$, which means that the risk of identifying a difference between the mean scores earned by the apprenticeship scheduler and the set of human performers when no such difference exists is less than $5\%$.

Results from a binomial test rejected the null hypothesis, indicating that the learned scheduling policy performed better than the human demonstrators in significantly more scenarios ($12$ versus $4$ scenarios; $p < 0.011$). In other words, we can say with $95\%$ certainty that the apprenticeship scheduler outperformed the average human player for the majority of the presented missile defense scenarios. This promising result was achieved using a relatively small training set, and suggests that learned policy can form the basis for a training tool to improve the average human player's score.  

\subsection{Real-World Data Set: Labor and Delivery}

Currently, nurse resource managers commonly operate without technological decision-making aids. As such, it is imprudent to introduce a fully autonomous solution for resource management, as doing so could have life-threatening consequences for practitioners unfamiliar with such automation. Rather, research has shown that a semi-autonomous system is preferable when integrating machines into human cognitive workflows ~\cite{kaber1997out,wickens2010stages}. Such a system would provide recommendations that a human supervisor could then accept or modify, and would be placed within the ``4-6" range on Sheridan's 10-point scale for levels of automation ~\cite{Parasuraman:2000}.

We found it prudent to test our apprenticeship scheduling technique with the algorithm offering recommendations to labor nurses who would evaluate how acceptable they found the quality of each recommendation. Specifically, we wanted to test whether the algorithm was able to learn to differentiate between high- and low-quality resource management decisions. If nurses accepted what the apprenticeship scheduler had learned to be high-quality advice while rejecting what the scheduler had learned to be low-quality advice, we could be reasonably confident that the apprentice scheduler had captured the desired resource management policy.

The first step, then, was to train a decision tree using the pairwise scheduling model based on the data set described in Section \ref{sec:NurseDataSet} of resource nurses' scheduling decisions. Recall that this data set consisted of the results of expert resource nurses playing the simulation for $2 \sfrac{1}{2}$ hours, simulating $60$ hours of elapsed time on a real labor floor, and yielding a data set of more than $3,013$ decisions. 

Second, we invited 15 labor nurses, none of whom were among those involved in training the algorithm, to play the same simulation used to collect the data (Figure \ref{fig:LDSimulation}). However, instead of purely soliciting decisions from the player, the simulation used the apprenticeship scheduling policy to offer recommendations about how to manage patients. Specifically, whenever a new patient arrived in the simulated waiting room, the apprenticeship scheduler would offer advice recommending 1) which of six wards to admit that patient to, 2) which bed within that ward to place that patient, and 3) which nurse should care for that patient. Nurses would then either accept the advice, automatically implementing the decision, or reject the advice and implement their own decisions.

In order to generate high-quality advice, the apprenticeship scheduler simply applied Equation \ref{eq:priorityFn}. To generate low-quality advice, the apprenticeship scheduler applied Equation \ref{eq:policyInv}, which changes the maximization to a minimization, as follows:
\begin{equation}
\tau_i^* = \argmin\limits_{\tau_i \in \boldsymbol{\tau}}\sum\limits_{\tau_x \in \boldsymbol{\tau}} f_{priority}(\tau_i,\tau_x)
\label{eq:policyInv}
\end{equation}
However, such a minimization could create a straw-man counterpoint to the high-quality advice, demonstrating only that the apprenticeship scheduler learned at least hard constraints (e.g., ``do not assign a patient to an occupied bed'') rather than a gradation over feasible actions (e.g., ``assign a less-busy nurse to a new patient rather than a busier nurse''). As such, we also used the apprenticeship scheduler to generate low-quality but feasible advice by only considering $\tau_i \in \boldsymbol{\tau}$ such that $\tau_i$ was feasible, as determined through a manually-encoded schedulability test.

For each of the 15 nurse players, we conducted two trials with the simulation offering advice. In one trial, the advice was high-quality; in the other, the simulation offered low-quality advice randomly chosen to be low-quality but feasible or low-quality and infeasible. We hypothesized that nurses would accept advice during the high-quality trials and reject advice during the low-quality trials (regardless of feasibility). Each simulation trial was randomly generated, with each player experiencing different scenarios with differing advice. On average, a nurse would receive $8.5$ recommendations per trial, resulting in a total of 256 recommendations across all nurses and trials. 

The nurses accepted high-quality advice 88.4\% of the time (114 of 129 high-quality recommendations), while rejecting low-quality advice 88.2\% of the time (112 of 127 low-quality recommendations), indicating that the apprenticeship scheduling technique is able to learn a high-quality model for resource management decision making in the context of labor and delivery. In other words, the apprenticeship scheduler was able to learn context-specific strategies for hospital resource allocation and apply them to make reasonable suggestions about which tasks to perform and when.

Anecdotally, some of the advice was not accepted for reasons that could be easily remedied: For example, upon initiation of the test, we were unaware that one room on the labor and delivery floor was unique because it uniquely contained cardiac monitoring equipment. As such, the algorithm did not know to reason about that feature and sometimes offered a recommendation that was feasible but less preferable for patients with cardiac-related comorbidities. It was not until later that we learned from the nurses about this particular feature. Such findings motivate the need for active learning for improved feature solicitation in future work. We also note that inter-operator agreement among nurse demonstrators is unlikely to be 100\%. For these reasons, we believe learning a policy that can generate advice validated to be correct nearly 90\% of the time is a favorable result.

\section{Model for Collaborative Optimization via Apprenticeship Scheduling}

Apprenticeship scheduling is designed to simply emulate human expert scheduling decisions; in this work, we also use the apprenticeship scheduler in conjunction with optimization to automatically and efficiently produce optimal solutions to challenging real-world scheduling problems. Our approach, called Collaborative Optimization via Apprenticeship Scheduling (COVAS), involves applying apprenticeship scheduling to generate a favorable (if suboptimal) initial solution to a new scheduling problem. To guarantee that the generated schedule is serviceable, we augment the apprenticeship scheduler to solve a constraint satisfaction problem, ensuring that the execution of each scheduling commitment does not directly result in infeasibility for the new problem. COVAS uses this initial solution to provide a tight bound on the value of the optimal solution, substantially improving the efficiency of a branch-and-bound search for an optimal schedule.

We show that COVAS is able to leverage good (but imperfect) human demonstrations to quickly produce globally optimal solutions. We also report that COVAS can transfer an apprenticeship scheduling policy learned for a small problem to optimally solve problems with twice as many variables as any shown during training, and produce an optimal solution an order of magnitude faster than mathematical optimization alone. Here, we provide an overview of the COVAS architecture and present its two components: the policy learning and optimization routines.

\subsection{COVAS Architecture}
\begin{figure}
	\center
	\includegraphics[width = 0.75\textwidth]{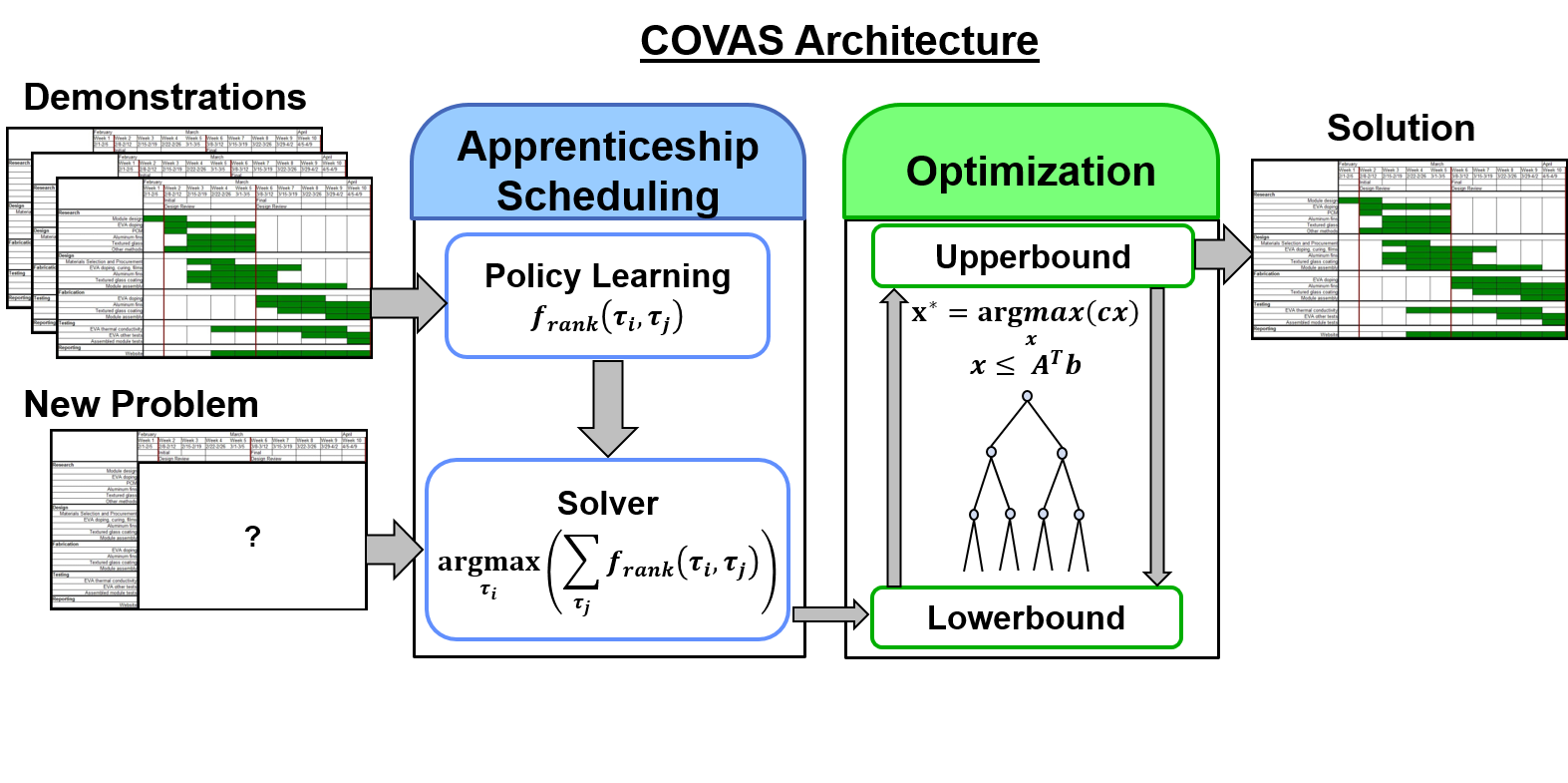}
	\caption{The COVAS architecture.}
	\label{fig:architecture}
\end{figure}
The system (Figure \ref{fig:architecture}) takes as input a set of domain expert scheduling demonstrations (e.g., Gantt charts) that contains information describing which agents complete which tasks, when and where. These demonstrations are passed to an apprenticeship scheduling algorithm that learns a classifier, {\normalsize$f_{priority}(\tau_i,\tau_j)$}, to predict whether the demonstrator(s) would have chosen scheduling action {\normalsize$\tau_i$} over action {\normalsize$\tau_j \in \boldsymbol{\tau}$}. Next, COVAS uses $f_{priority}(\tau_i,\tau_j)$ to construct a schedule for a new problem. The system creates an event-based simulation of this new problem and runs this simulation in time until all tasks have been completed. In order to complete tasks, COVAS uses $f_{priority}(\tau_i,\tau_j)$ at each moment in time to select the best scheduling action to take. We describe this process in detail in the next section. COVAS then provides this output as an initial seed solution to an optimization subroutine (i.e., a MILP solver). The initial solution produced by the apprenticeship scheduler improves the efficiency of a search by providing a bound on the objective function value of the optimal schedule. This bound informs a branch-and-bound search over the integer variables \cite{Bertsimas:2005}, enabling the search algorithm to prune areas of the search tree and focus its search on areas that can yield the optimal solution. After the algorithm has identified an upper- and lowerbound within some threshold, COVAS returns the solutions that have proven optimal within that threshold. Thus, an operator can use COVAS as an anytime algorithm and terminate the optimization upon finding a solution that is acceptable within a provable bound.

\subsection{Apprenticeship Scheduling Subroutine}
In Section \ref{sec:apprenticeshipScheduling}, we presented our apprenticeship scheduling algorithm, which is centered around learning a classifier, $f_{priority}(\tau_i,\tau_j)$, to predict whether an expert would take scheduling action $\tau_i$ over $\tau_j$. With this function, we can then predict which single action $\tau_i^*$ amongst a set of actions $\boldsymbol{\tau}$ the expert would take by applying Equation \ref{eq:priorityFn}.
In this section, we build upon this formulation and integrate it into our collaborative-optimization via apprenticeship scheduling framework.

As a subroutine within COVAS, $f_{priority}(\tau_i,\tau_j)$ is applied to obtain the initial solution to a new scheduling problem as follows: First, the user must instantiate a simulation of the scheduling domain; then, at each time step in the simulation, take the scheduling action predicted by Equation \ref{eq:priorityFn} to be the action that the human demonstrators would take. This equation identifies the task $\tau_i$ with the highest importance marginalized over all other tasks {\normalsize$\tau_j \in \boldsymbol{\tau}$}.
Unlike our original formulation in Section \ref{sec:apprenticeshipScheduling}, each selected action is validated using a schedulability test (i.e., solving a constraint satisfaction problem) to ensure that direct application of that action does not violate the constraints of the new problem. For example, in anti-ship missile defense, one would check to ensure that the given action does not result in a suicidal deployment (i.e., the decoy directly causes a missile to impact the ship). This test must be fast, so as to make the benefit to feasibility and optimality in the resulting schedule worth the additional complexity. If, at a given time step, {\normalsize$\tau_i^*$} does not pass the schedulability test, COVAS uses Equation \ref{eq:priorityFn} for all {\normalsize$\tau_i \in \boldsymbol{\tau}\backslash \tau_i^*$} to consider the second-best action. If no action passes the schedulability test, no action is taken during that time step.

While the schedulability test forces the apprenticeship scheduling algorithm to follow a subset of the full constraints in the MILP formulation, it is possible that the algorithm may not successfully complete all tasks. Here, we model tasks as optional and use the objective function to maximize the total number of tasks completed. In turn, constraints for a task that the apprenticeship scheduling algorithm did not satisfactorily complete can be turned off, with a corresponding penalty in the objective function score. Thus, an initial seed solution that has not completed all tasks (i.e., satisfied all constraints to complete the task) can still be helpful for seeding the MILP.

\subsection{Optimization Subroutine}
For optimization, we employ mathematical programming techniques to solve mixed-integer linear programs via branch-and-bound search. COVAS incorporates the solution produced by the apprenticeship scheduler to seed a mathematical programming solver with an initial solution, which is a built-in capability provided by many off-the-shelf, state-of-the-art MILP solvers, including CPLEX\footnote{IBM ILOG CPLEX Optimization Studio {http://www-03.ibm.com/software/products/en/ibmilogcpleoptistud}} and Gurobi\footnote{Gurobi Optimization, Inc. {http://www.gurobi.com}}. This seed provides a tight bound on the objective function value of the optimal solution, which serves cut the search space; these cuts allow COVAS to more quickly hone in on the optimal solution. Furthermore, this approach allows COVAS to quickly achieve a bound on the optimality of the solution provided by the apprenticeship scheduling subroutine. In such a manner, an operator can determine whether the apprenticeship scheduling solution is acceptable or whether waiting for successive solutions from COVAS is warranted.

\section{Results and Discussion}
In this section, we empirically validate that COVAS is able to generate optimal solutions more efficiently than state-of-the-art optimization techniques. \color{black} We also analyze the sensitivity of the computational time COVAS required to find an optimal solution as a function of the quality of the scheduling policy learned by the apprenticeship scheduling algorithm. \color{black}

\subsection{Validation Against Expert Benchmark}

\begin{figure}
	\center
	\includegraphics[width = 0.5\textwidth]{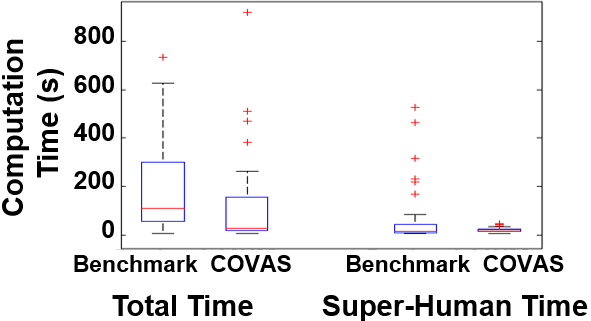}
	\caption{The total computation time for COVAS, as well as the time COVAS required to identify a solution superior to that resulting from a human expert's demonstration. \color{black}Results for the benchmark and COVAS are depicted offset to the left and right of each position along the x-axis, respectively. \color{black}}
	\label{fig:CombinedDeterministic}
\end{figure}

In this section, we empirically validate that COVAS is able to generate optimal solutions more efficiently than state-of-the-art optimization techniques. As a baseline benchmark, we solve a pure MILP formulation (Appendix \ref{sec:AppendixASMD} Equations \ref{eq:objFunc}-\ref{eq:trackingShipGEQ}) using Gurobi, which applies state-of-the-art techniques for heuristic upperbounds, cutting planes and LP relaxation lowerbounds. We set the optimality threshold at $10^{-3}$. For the apprenticeship scheduling subroutine's schedulability test, we apply Equations \ref{eq:suicide1}-\ref{eq:suicide2} as a constraint satisfaction check when testing the feasibility of action {\normalsize$\tau_i^*$}, given by applying Equation \ref{eq:priorityFn}. With regard to tasks within the apprenticeship scheduler's seed solution that are not satisfactorily completed, the MILP can leave those tasks incomplete to start by initially setting {\normalsize$V_{m} \leftarrow 0$}.

We trained COVAS' apprenticeship scheduling algorithm on demonstrations of experts' solutions to unique ASMD scenarios (save for one ``hold-out" scenario) from the ASMD data set described in Section \ref{sec:ASMDDataSet}. We then tested COVAS on the hold-out scenario. We also applied a pure MILP benchmark on this scenario and compared the performance of COVAS to the benchmark. We generated one data point for each unique demonstrated scenario (i.e., leave-one-out cross-validation) to validate the benefit of COVAS.

Figure \ref{fig:CombinedDeterministic} consists of two performance indicators: The total computation time required for the MILP benchmark and COVAS to solve for the optimal solution is depicted on the left; to the right is the computation time required for the benchmark and COVAS to identify a solution better than that provided by a human expert. This figure indicates that COVAS was not only able to improve overall optimization time, but that it also substantially improved computation time for solutions superior to those produced by human experts. The average improvements in computation time with COVAS were $6.7$x the overall optimization time and $3.1$x the expert-generated solutions.

\begin{figure}
	\center
	\includegraphics[width = 0.5\textwidth]{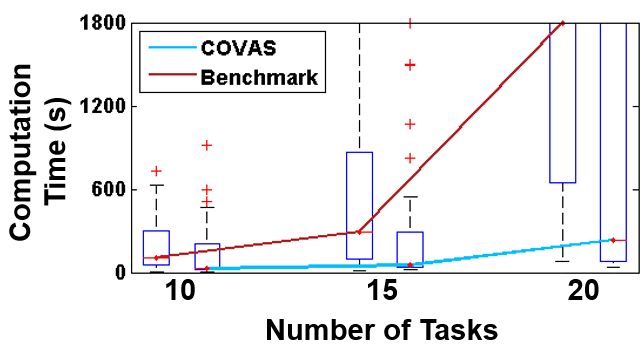}
	\caption{The total computation time needed for COVAS and the MILP benchmark to identify the optimal solution for the tested scenarios. \color{black}Results for the benchmark and COVAS are depicted offset to the left and right of each position along the x-axis, respectively. \color{black}}
	\label{fig:TransferLearning}
\end{figure}

Next, we evaluated COVAS' ability to transfer prior learning to more-challenging task sets. We trained on a level in the ASMD game in which a total of 10 missiles of varying types came from specific bearings at given times. We randomly generated a set of scenarios involving 15 and 20 missiles, with bearings and times randomly sampled with replication from the set of bearings used in the 10-missile scenario.

Figure \ref{fig:TransferLearning} depicts the computation time required by COVAS and the MILP benchmark to identify the optimal solution for scenarios involving 10, 15 and 20 missiles. The average improvement to computation time with COVAS was $4.6$x, $7.9$x, and $9.5$x, respectively, demonstrating that COVAS is able to efficiently leverage the solutions of human domain experts to quickly solve problems twice as large as those the demonstrator provided for training.

\subsection{Sensitivity Analysis of COVAS to Apprenticeship Scheduler's Learned Policy}
\label{sec:Advantage}
\color{black}
Here, we assess the sensitivity of the computational time COVAS required to find an optimal solution as a function of the quality of the scheduling policy learned by the apprenticeship scheduling algorithm.

\color{black}
\subsubsection{Sensitivity Analysis Design}
We sought to understand how incorrect predictions generated by the apprenticeship scheduling algorithm's classifier, $f_{priority}(.,.)$, would affect COVAS' computational efficiency. We considered three classes of mistakes that the apprenticeship scheduler could make when creating an initial schedule: two types of mistakes related to agent allocation (swapping tasks among agents and the misallocation of agents to a particular task), as well as task sequencing errors. We generated a synthetic dataset involving these three error classes as follows:

\color{black}
\begin{itemize}
	\item {\emph{Allocation: Swapping}: Select two tasks with uniform probability, $\tau_i$ and $\tau_j$, such that the agent $a$ assigned to $\tau_i$ is different from the agent $a'$ assigned to $\tau_j$, and subsequently swap their assignment such that agent $a'$ now performs $\tau_i$ and vice-versa.  }
	\item {\emph{Allocation: Stealing}: Select one task, $\tau_i$, \color{black}with uniform probability, where $\tau_i$ is\color{black}~assigned to agent $a$, and reassign it to a different agent, $a'$.}
	\item {\emph{Sequencing}: Select two tasks, $\tau_i$ and $\tau_j$, \color{black}with uniform probability,\color{black}~ such that $\tau_i$ precedes $\tau_j$ in the schedule, and reverse their order such that $\tau_j$ now precedes $\tau_i$.}
\end{itemize}

\color{black}
Table \ref{tab:expDesign} depicts the experiment design for our sensitivity analysis. We incorporated the synthetic dataset because the scheduling problem has a well-defined objective function and set of constraints for use in the optimization component of COVAS, and also because the data set encompasses three different types of scheduling problems. The problem domain and the mock demonstrator's heuristics for each problem were defined in Section \ref{sec:syntheticDataSet}. We generated 15 problems for each problem type, misclassification type, and number of misclassifications. Five replicates were generated for each problem, with the replicates varying according to misclassification type (e.g., switching the ordering of $\tau_i$ and $\tau_j$ versus switching $\tau_p$ and $\tau_q$). In total, the analysis involved $3\times3\times3 =27$ different experimental settings and $27\times15\times5=2,025$ total data points.

\color{black}
\begin{table}[b]
	\begin{center}
		\begin{tabular}{ |c||c|c|c|c|c|c|c|c|c|c|c| } \hline
			\multirow{2}{*}{Problem Type} & \multicolumn{9}{c|}{Travel} & Resource & Temporal \\
			& \multicolumn{9}{c|}{Distance} & Contention & Requirements \\
			(Heuristic Applied) & \multicolumn{9}{c|}{(Equation \ref{eq:VRPRule})} & (Eq. \ref{eq:RCRule}) & (Eq. \ref{eq:EDFRule}) \\ \hline
			Misclassification Type & \multicolumn{3}{c|}{Swapping} & \multicolumn{3}{c|}{Stealing} & \multicolumn{3}{c|}{Sequencing} & $\ldots$ & $\ldots$  \\ \hline
			\# Misclassifications & $1$ & $2$ & $3$ & $1$ & $2$ & $3$ & $1$ & $2$ & $3$ & $\ldots$ & $\ldots$  \\ \hline
		\end{tabular}
	\end{center}
	\caption{\color{black}This table depicts the experimental design for COVAS' sensitivity analysis.\color{black}}
	\label{tab:expDesign}
\end{table}

\subsubsection{Statistical Model for the Analysis}

\color{black}
We performed a mixed-effects multiple linear regression to quantify the sensitivity of COVAS with respect to the quality of the apprenticeship scheduling policy. The dependent variable was the \emph{computational time} required by COVAS to identify the optimal solution. The independent variables were the \emph{problem type} (vehicle routing, resource contention, and temporal requirements), the \emph{misclassification type} (allocation-based swapping and stealing and sequencing-based errors), the \emph{number of errors} (one, two, or three), and the \emph{objective function value} of the schedule produced by the apprenticeship scheduling algorithm (normalized to the objective function value of the optimal solution). The aforementioned independent variables were modeled as fixed effects. We also included random effects for the individual problem and for the optimality of the apprenticeship scheduler's solution as a function of the fixed-effects independent variables. We applied a Box-Cox transformation~\cite{box1964analysis} to normalize the data for regression, which returned $\lambda = 0.223$ as the optimal transformation factor. We established statistical significance for the regression parameters at the $\alpha = 0.05$ level.


\subsubsection{Results and Discussion}

\begin{table}
	\begin{center}
		\begin{tabular}{ l|l|c|c|c } 
			\multicolumn{2}{l|}{Parameter} & Estimate & Confidence Interval &  p-value\\ \hline\hline
			\multicolumn{2}{l|}{Intercept} & $-13.095$ & $(-15.047,-11.143)$ & $\mathbf{<0.001}$ \\ \hline
			\multicolumn{2}{l|}{Obj. Fn. Val. of AS's schedule (Normalized)}	& $5.256$ & $(3.937,6.575)$ & $\mathbf{<0.001}$\\ \hline 
			\multicolumn{2}{l|}{Number of Classification Errors} & $1.357$ & $(0.630,2.083)$ & $\mathbf{<0.001}$\\ \hline 
			Allocation vs. Sequencing & Allocation & $6.471$ & $(4.942,7.999)$ & $\mathbf{<0.001}$\\ \cline{2-5}
			Misclassification & Sequencing & - & - & - \\ \hline 			
			Allocation  & Steal & $1.018$ & $(-0.448,2.484)$ & $0.173$\\ \cline{2-5}
			Misclassification Type & Swap & - & - & - \\ \hline 
			\multirow{3}{*}{Problem Type:}&Temporal Req. 		& $3.074$ & $(1.300,4.848)$ & $\mathbf{<0.001}$\\ \cline{2-5}
			& Resource Contention & $0.271$ & $(-1.893,2.435)$ & $0.806$\\ \cline{2-5}
			&Travel Distance & - & - & - \\ \hline \hline			
		\end{tabular}
	\end{center}
	\caption{\color{black}This table depicts the results of the regression analysis. Entries with dashes indicate that the associated parameter setting was the baseline. Statistically significant values are in bold.\color{black}}
	\label{tab:ANOVA}
\end{table}

\color{black}
Table \ref{tab:ANOVA} reports the statistical results of our sensitivity analysis.\footnote{The Akaike information criterion~\cite{akaike1974new} and Bayesian information criterion~\cite{schwarz1978estimating} are 2,968.1 and 3,016, respectively. The log likelihood of the model is -1,474.1, and the deviance is 2,948.1.} There are three key takeaways from these results: First, the primary driver of COVAS' computational time was the objective function value of the schedule produced by the apprentice scheduler's learned policy -- not the number of misclassification errors made when constructing the schedule. As shown in the third and fourth rows of Table \ref{tab:ANOVA}, the impact of the objective function value on COVAS' computational efficiency was $5.256/1.357\approx 4$ times more than for the individual classification errors made by $f_{priority}(.,.)$.


Second, there was a statistically significant effect for allocation- versus sequencing-based perturbations. The regression analysis shows that COVAS's improvement in computation time lessens when considering allocation-based perturbations ($p < 0.001)$. However, there was no statistically significant effect present between allocation-based swapping and stealing errors ($p = 0.173$). 

Third, we observed a sensitivity to the problem type / heuristic applied, with the problem type emphasizing temporal requirements (for which the heuristic in Equation \ref{eq:EDFRule} is applied) representing the most challenging problem. There was not a significant difference between the resource contention and VRP-style problem types ($p = 0.806$).

Finally, note that COVAS showed an improvement in computation time relative to a commercial, state-of-the-art solver regardless of problem and misclassification type and number. We performed separate regression analyses for each problem type and as a function of whether the classification errors by the apprenticeship scheduler were allocation- or sequencing-based. Table \ref{tab:Individual} depicts the objective function value of the apprenticeship scheduler's schedule for which COVAS no longer demonstrates an advantage in computation time (relative to a state-of-the-art solver) when marginalizing over the number of errors. For example, if COVAS is scheduling a problem for which travel distance is key (and the apprentice scheduler was trained on a mock demonstrator applying Equation \ref{eq:VRPRule}) and the apprentice scheduler makes a number of (1, 2, or 3) allocation-based swapping-type classification errors while constructing the schedule, COVAS is faster than a state-of-the-art benchmark, so long as the objective function value of the schedule produced by the apprentice scheduler is no worse than 1.88 times that of the optimal solution. The results show that COVAS demonstrates an advantage for all problem and misclassification types. 

\begin{table}
	\begin{center}
		\begin{tabular}{ l|c|c|c } 
			& Travel  & Resource  & Temporal  \\ 
			&  Distance &  Contention &  Requirements \\ \hline \hline			
			Allocation-Based Swapping Errors & 1.68 & 1.38 & 1.13 \\ \hline
			Allocation-Based Stealing Errors & 1.88 & 1.43 & 1.14 \\ \hline
			Sequencing-Based Errors & 1.34 & 1.27 & 1.12 \\ \hline
		\end{tabular}
	\end{center}
	\caption{\color{black}This table depicts the maximum objective function value of the apprenticeship scheduler's initial solution (normalized to that of the optimal solution) to provide COVAS' optimization subroutine with an improvementin computation time.\color{black}}
	\label{tab:Individual}
\end{table}

\subsubsection{Conclusion}

The results from our sensitivity analysis support the hypothesis that COVAS is robust to misclassification errors by the apprenticeship scheduler's learned policy. The data indicate that the apprenticeship scheduler's solution quality is the dominating factor, rather than the number of individual mistakes made when generating that solution. \color{black} While further investigation of other scheduling problems is warranted, the variants within this data set inform our understanding of the sensitivity of COVAS to imperfections in the learned policy across a range of problem types. 

These results also provide insight into ways that we can potentially improve COVAS' apprenticeship scheduling subroutine. For example, COVAS is more sensitive to the objective function value of the schedule produced by the apprenticeship scheduler's policy, while being somewhat robust to the number of errors made by the apprenticeship scheduler when constructing the schedule. As such, imitation learning~\cite{ross2011reduction,cheng2018convergence} approaches, which attempt to bootstrap off of an initial, learned policy, may be able to improve the quality of the solutions produced by the apprenticeship scheduler. Further, a Bayesian IRL approach~\cite{Michini:2012,Ramachandran:2007}, which seeks to infer a policy mimicking an ``ideal demonstrator,'' may also be able to leverage demonstrations to better guide COVAS' optimization subroutine. In future work, we will seek to determine how to combine such approaches with our pairwise training procedure for COVAS' apprenticeship scheduling subroutine.

\color{black}

\section{Limitations and Future Work}

The core of the apprenticeship scheduling algorithm is learning a classifier, $f_{priority}(\tau_i,\tau_j)$, to predict whether a human expert would take action $\tau_i$ over $\tau_j$. The output of $f_{priority}(\tau_i,\tau_j)$ is a probability in $[0,1]$. This pairwise approach has a number of key advantages: For example, it is nonparametric with regard to the number of tasks, meaning one can learn from problems involving $n$ actions and apply that knowledge to problems with $n' \neq n$ actions. However, there are two interesting anomalies inherent in this approach: First, one could hypothetically evaluate $f_{priority}(\tau_i,\tau_j)$ and find that it predicts that the expert has a higher probability of taking action $\tau_i$ than $\tau_j$; however, evaluating $\argmax\limits_{\tau_i \in \boldsymbol{\tau}} \sum\limits_{\tau_j \in \boldsymbol{\tau}} f_{priority}(\tau_i,\tau_j)$ could predict that $\tau_j$ is the action most likely to be taken by the expert. The second anomaly entails the lack of a guarantee that the transitive property will hold for arbitrary $f_{priority}(\tau_i,\tau_j)$. For example, it could be that $f_{priority}(\tau_i,\tau_j) > 0.5$, $f_{priority}(\tau_j,\tau_k) > 0.5$, but also $f_{priority}(\tau_k,\tau_i) > 0.5$ for some $\tau_i$, $\tau_j$, and $\tau_k$.  Through our evaluation, we have shown that the formulation for apprenticeship scheduling can learn high-quality policies from human domain experts' demonstrations. However, an interesting aim for future work would be to study these anomalies, quantify their effects -- if any -- and develop a formulation to alter these effects. Appendix \ref{sec:AppendixTriangle} provides an example formulation for how to mitigate such anomalies.

COVAS also has some interesting aspects that merit future investigation. COVAS is able to leverage expert scheduling demonstrations to speed up the computation of provable, globally optimal scheduling solutions. However, the approach is still limited by the quality of the demonstrations provided by experts, as well as the ability of the apprenticeship scheduling algorithm to generalize the information within those demonstrations. The MILP's computation time is expedited by tight upperbounds (i.e., an initial seed) provided by the apprenticeship scheduling algorithm. If the apprenticeship scheduling algorithm is unable to provide a tight upperbound, the MILP's computation time may not be significantly improved. In future work, we will explore potential extensions to the apprenticeship scheduling algorithm to improve its ability to learn from noisy demonstrations. One approach could be to incorporate a trustworthiness metric \`a la \color{black}\citeA{Zhang:2009}\color{black}~directly into the training of the classifier to uncover a latent action ranking. For example, instead of binary labels, we could reformulate the problem to be one of regression, where positive and negative labels are proportional and inversely proportional, respectively, to the fidelity of the demonstrator. 

\color{black}
Finally, apprenticeship scheduling with heterogeneous demonstrators is an important area for future work. In this paper, we demonstrated the ability to learn from 1) homogeneous demonstrators with varying quality and quantity of training data in a synthetic domain and 2) heterogeneous demonstrators in multiple real-world domains (i.e., healthcare and ship defense). Future lines of research include learning clusters of operator archetypes through unsupervised or semi-supervised learning so that the apprenticeship scheduler can better account for individual differences between operators. If each demonstrator applies different strategies, it will be more difficult to generalize across operators~\cite{Sammut:1992}. However, if there are a small number of demonstrator types relative to the number of demonstrators, it may be possible to leverage commonality within or across types to bootstrap the learning process.
\color{black}

\section{Conclusions}
In this paper, we proposed a technique for apprenticeship scheduling that relies upon a pairwise comparison of scheduled and unscheduled tasks to learn a model for task prioritization. We validated that our apprenticeship scheduling algorithm is able to learn high-quality scheduling policies from demonstration across both synthetic data and real-world data sets. Specifically, apprenticeship scheduling can learn from nurse resource managers to make scheduling decisions that are accepted by resource nurses $90\%$ of the time, and can learn from military experts to solve a variant of the weapon-to-target assignment problems with better performance than the average human expert. Next, we embedded this apprenticeship scheduling algorithm within a ML-optimization framework. This algorithm, COVAS, leverages the ability of apprenticeship scheduling to capture the knowledge of human domain experts in order to produce optimal solutions for complex real-world scheduling problems. We validated our technique using a data set collected from human experts solving an anti-ship missile defense problem, and showed that our approach can substantially improve upon solutions produced by experts, at a rate up to $9.5$ times faster than an optimization approach that does not incorporate human expert demonstration.

\balance

\vskip 0.2in
\bibliography{paper}
\bibliographystyle{theapa}

\clearpage

\appendix

\section{ASMD: Mathematical Program Formulation}
\label{sec:AppendixASMD}
We readily formulate the ASMD as a mixed-integer linear program in Equations \ref{eq:objFunc}-\ref{eq:trackingShipGEQ}. This formulation incorporates a set of binary decision variables: {\normalsize$A_{d,m,t}\in\{0,1\}$} is set to 1 to indicate that decoy $d$ is assigned to missile $m$ at time $t$, and is 0 otherwise. {\normalsize$A_{d,t}\in\{0,1\}$} is set to 1 to indicate that decoy $d$ is assigned to some missile at time $t$, and is 0 otherwise. {\normalsize$U_{d,m} \in \{0,1\}$} is set to 1 to indicate that decoy $d$ is used against missile $m$, and is 0 otherwise. {\normalsize$U_d \in \{0,1\}$} is set to 1 to indicate that decoy $d$ is used in the solution, and is 0 otherwise. {\normalsize$X_{d,l}\in \{0,1\}$} is set to 1 to indicate that decoy $d$ is deployed at location $l$, and is 0 otherwise. {\normalsize$V_m\in\{0,1\}$} is set to 1 to indicate that missile $m$ has been effectively diverted, and is 0 otherwise. {\normalsize$G_{g,m,t}\in\{0,1\}$} is set to 1 to indicate that missile $m$ is tracking the ship at time $t$. A single missile might have multiple, separate epochs during which it tracks the ship (e.g., it first tracks the ship, then a decoy, then the ship again after the decoy disappears); thus, the program can choose which index $g$ to represent the various epochs in {\normalsize$G_{g,m,t}$}. {\normalsize$J_{d,m}\in\{0,1\}$} is set to 1 to indicate that decoy $d$ is deployed after missile $m$'s flight (i.e., after it either hits the ship or is guided astray by a decoy).

The program contains the following continuous variables: {\normalsize$S^{decoy}_{d,m}$} represents the start time of the assignment of decoy $d$ to missile $m$, and {\normalsize$S^{decoy}_{d}$} is the time at which decoy $d$ is deployed from the ship. Likewise, {\normalsize$F^{decoy}_{d,m}$} represents the finish time of the assignment of decoy $d$ to missile $m$, and {\normalsize$F^{decoy}_{d}$} is either the time at which the decoy disappears or the end of the engagement. {\normalsize$S^{ship}_{g,m}$} indicates the start time of missile $m$ tracking the ship during epoch $g$, and {\normalsize$F^{ship}_{g,m}$} indicates the finish time of missile $m$ tracking the ship during epoch $g$. We include the constant, $M$, which is a large, positive number allowing one to formulate linear, conditional constraints.

The program also includes the following set of constants: {\normalsize$dt^{re-target}_m$} is the length of time for which a missile will track a single target (i.e., a decoy or ship) before reassessing which target is best to track. Thus, if the missile begins tracking the ship at time $t$, no decoy can break its lock during the interval {\normalsize$[t,t+dt^{re-target}_m)$}. {\normalsize$ETA_m$} is the time at which missile $m$ will reach the ship's immediate vicinity. {\normalsize$t^{appear}_m$} is the time at which missile $m$ first becomes close enough to track the ship. {\normalsize$c_d$} represents the financial cost of deploying decoy $d$. $\alpha,\alpha',$ and $\alpha''$ are predefined weighting terms for the objective function. The computational complexity of completely searching for the optimal solution via this formulation is dominated by the integer variables, which yields {\normalsize$O(2^{dmt + dm + dt + dl + d + gmt + m})$}.

Equation \ref{eq:objFunc} is a multi-criteria objective function that minimizes a weighted linear combination of the cost of all decoy deployments, less the total time during which missiles are tracking decoys and the number of missiles successfully guided away from the ship. Equations \ref{eq:AdmtAdt}-\ref{eq:FSAdmtUdm} ensure internal consistency between the variables. Equation \ref{eq:EvapTime} ensures that a decoy, if deployed, is active for $dt_{d}^{evap}$ units of time given its timing characteristics. Equation \ref{eq:Xdl} ensures that a decoy is deployed to no more than one location. Equation \ref{eq:GoodPos} ensures that, if a decoy is deployed against a missile, its deployment location will be a more attractive target than the ship for that missile. Equation \ref{eq:AdmtGgmt} requires that each missile track either a ship or decoy while within range. Equations \ref{eq:suicide1}-\ref{eq:suicide2} force a decoy, if deployed to a location that would cause missile $m$ to impact the ship, to either be deployed after the missile has already been diverted or reached the ship (Equation \ref{eq:suicide1}) or to be deployed and disappear before the missile enters targeting range (Equation \ref{eq:suicide2}).

Equation \ref{eq:Vm} ensures that a missile must be tracking a decoy in the final seconds before it reaches the vicinity of the ship, or else the missile will impact the ship. The duration of this critical period is dependent upon missile dynamics and the target selection process. Equation \ref{eq:BetterLocation} ensures that a missile will select the most attractive decoy according to that missile's selection logic. Equation \ref{eq:Sweeping} restricts decoy deployments such that the missile heading does not ``sweep" across the ship in the final seconds of the missile's flight. If a missile does not have enough time to change its direction toward a newly deployed decoy, that missile will fly into the ship.

Equations \ref{eq:SGgmt}-\ref{eq:trackingShipGEQ} ensure that the duration of epoch $g$ of missile $m$ while tracking the ship lasts exactly as long as the retargeting time for the missile. Equations \ref{eq:SGgmt}-\ref{eq:GgmtF} are akin to Equations \ref{eq:SdmAdmt}-\ref{eq:FdmAdmt} and relate the start and finish times of ship-tracking epoch $g$ to the decision variable {\normalsize$G_{g,m,t}$}.  Equation \ref{eq:FSGgmt} is akin to Equation \ref{eq:FSAdmtUdm} and relates the start and finish times of ship-tracking epoch $g$ to the decision variable {\normalsize$G_{g,m,t}$}. Equation \ref{eq:trackingShipGEQ} ensures that the tracking time is {\normalsize$dt_{m}^{re-target}$} if the missile is airborne for at least {\normalsize$dt_{m}^{re-target}$} seconds. Otherwise, the tracking time is equal to the time before impacting the ship (i.e., {\normalsize$ETA^m-t-1$}). Finally, a term (i.e., {\normalsize$-MG_{g,m,t-1}$}) disables the constraint for all $t$ except for the exact moment when $t$ begins tracking the ship.
\begin{align}
	\min z \text{, } z&=  \alpha\sum_{d}c_dU_d-\alpha'\sum_{d,m,t} A_{d,m,t}-\alpha''\sum_{m}V_m \label{eq:objFunc} \\
	A_{d,m,t} &\leq A_{d,t}, \forall d,m,t \label{eq:AdmtAdt}\\
	A_{d,m,t} &\leq U_{d,m},\forall d,m,t \label{eq:AdmtUdm}\\
	X_{d,l} &\leq U_d , \forall d,l \label{eq:XdlUd}\\
	S^{decoy}_{d} &\leq S^{decoy}_{d,m} , \forall d,m \label{eq:SS}\\
	S^{decoy}_{d,m} &\leq t + M(1-A_{d,m,t}), \forall d,m,t \label{eq:SdmAdmt}\\
	F^{decoy}_{d,m} &\leq F^{decoy}_{d} , \forall d,m \label{eq:FF}\\
	tA_{d,m,t} &\leq F^{decoy}_{d,m}, \forall d,m,t \label{eq:FdmAdmt}\\
	M(U_{d,m}-1) & \leq S^{decoy}_{d,m}-F^{decoy}_{d,m}-1 + \sum_t A_{d,m,t} \leq M(1-U_{d,m}) \label{eq:FSAdmtUdm}\\
	M(U_d-1)  &\leq F^{decoy}_{d} - S^{decoy}_{d} -  dt^{evap}_d \leq M(1-U_d) \label{eq:EvapTime}\\
	\sum_l X_{d,l} &\leq 1 , \forall d \label{eq:Xdl}\\
	U_{d,m} &\leq \sum_{ l | m \text{ seduced by decoy } d \text{ in location } l} X_{d,l} ,\forall d,m \label{eq:GoodPos}\\
	1 &= \sum_d A_{d,m,t} + \sum_g G_{g,m,t}, \forall m,t \label{eq:AdmtGgmt}
\end{align}
\begin{gather}
	t_m^{appear}-F^{decoy}_d \geq M(X_{d,l}+V_m-J_{d,m}-2), \nonumber\\ \forall d,l,m \text{ s.t. decoy d in location l would cause missile $m$ to impact the ship.}
	\label{eq:suicide1}
\end{gather}
\begin{gather}
	S^{decoy}_d - ETA_m \geq M(X_{d,l}+V_m+J_{d,m}-3), \forall d,l,m \text{ s.t. decoy d} \nonumber\\ \text{in location l would cause missile $m$ to impact the ship.}
	\label{eq:suicide2}
\end{gather}
\begin{gather}
	V_m \leq \sum_d A_{d,m,t}, \forall m,t | t \text{ in critical region for missile } m \text{.}
	\label{eq:Vm}
\end{gather}
\begin{gather}
	2 \geq A_{d,m,t} + X_{d,l} + X_{d',l'} ,  \forall d,d',l,l',m,t  \text{ s.t. missile m is more} \nonumber \\ \text{attracted to decoy d' at location l' than decoy d at location l at time t.} \label{eq:BetterLocation}
\end{gather}
\begin{gather}
	1 \geq A_{d,m,t} + A_{d',m,t},\forall d,d',m,t  \text{ s.t. } d\neq d'  \nonumber\\
	\text{ and t is in a critical region before impact.} \label{eq:Sweeping}
\end{gather}
\begin{align}
	S^{ship}_{g,m} &\leq t + M(1-G_{g,m,t}), \forall g,m,t \label{eq:SGgmt}\\
	t*G_{g,m,t} &\leq F^{ship}_{g,m}, \forall g,m,t  \label{eq:GgmtF} \\
	M(U_{g,m}-1) & \leq S^{ship}_{g,m}-F^{ship}_{g,m}-1 + \sum_t G_{g,m,t} \leq M(1-U_{g,m}) \label{eq:FSGgmt} \\
	F^{ship}_{g,m} - S^{ship}_{g,m}  &\geq M(G_{g,m,t}-1)  \nonumber \\
	&+\begin{cases}
		dt^{re-target}_m-1 &\text{if } t < ETA_m- dt^{re-target}_m,\\
		ETA_m-t-1 &\text{ otherwise.}
	\end{cases} \nonumber \\
	&+\begin{cases}
		-MG_{g,m,t-1} &\text{if } t > t^{appear}_m,\\
		0 &\text{ otherwise.}
	\end{cases}
	\nonumber \\
	&\forall g,m,t | t^{appear}_m \leq t < ETA_m
	\label{eq:trackingShipGEQ}
\end{align}

As ASMD is a time-extended problem, the formulation must discretize time. However, note that the granularity with which the task of protecting the ship is decomposed as a function of time is a modeling choice with ramifications for the quality and computation time of a solution. Consider a missile that will hit the ship if it tracks the ship in some time interval $[t,t')$ for a duration $dt = t-t'$. 
The captain might, at time $t$, deploy a decoy $d$, such as a hovering UAV, that is able to last the entire duration $dt$. However, it may be preferable to deploy one or more decoys, $d'$, each of which remains active for a portion of the specified time interval. Furthermore, in a situation wherein another missile, $m'$, is launched before $m$, it may be best to have a decoy deployed before $t$ that can divert both $m$ and $m'$ during part or all of those missiles' flights.

As we do not know a priori the best time to deploy a decoy that can be used for varying portions (i.e., subtasks) of the task of mitigating each missile, we must decompose the task into sufficiently small time steps. Discretizing time exponentially increases the search space, and thus the time to compute the solution; therefore, there is a balance between optimality (and feasibility) and computation time. In order to generate an exact solution, we chose the least-common multiple of the time constants, which is trivially $1$, as the unit of time in the simulation.

\section{Mitigating Anomalies}
\label{sec:AppendixTriangle}
To mitigate anomalies inherent in a pairwise comparison approach, one could consider the following formulation in Equations \ref{eq:opTree1} through \ref{eq:opTree6} when learning a decision tree model, $T^*$, for apprenticeship scheduling:

\begin{equation}
T^* = \argmin\limits_{T} \mathbb{E}_{\theta,y}[L(y^m_{\tuple{\tau_i,\tau_j}},T(^{rank}\theta^m_{\tuple{\tau_i,\tau_j}}))]
\label{eq:opTree1}
\end{equation}
subject to
\begin{align}
	T(^{rank}\theta^m_{\tuple{\tau_i,\tau_j}}) & > 0.5 + M(1-Z_{i,j}), \forall \tau_i,\tau_j \label{eq:opTree2}\\
	T(^{rank}\theta^m_{\tuple{\tau_i,\tau_j}}) & < 0.5 + M(Z_{i,j}), \forall \tau_i,\tau_j \label{eq:opTree3}\\
	\sum\limits_{\tau_k \in \boldsymbol{\tau}} T(^{rank}\theta^m_{\tuple{\tau_i,\tau_k}})
	- \sum\limits_{\tau_k \in \boldsymbol{\tau}} T(^{rank}\theta^m_{\tuple{\tau_j,\tau_k}}) &> M(1-Z_{i,j}), \forall \tau_i,\tau_j \label{eq:opTree4} \\
	\sum\limits_{\tau_k \in \boldsymbol{\tau}} T(^{rank}\theta^m_{\tuple{\tau_i,\tau_k}})
	- \sum\limits_{\tau_k \in \boldsymbol{\tau}}T(^{rank}\theta^m_{\tuple{\tau_j,\tau_k}}) &< M(Z_{i,j}), \forall \tau_i,\tau_j \label{eq:opTree5} \\
	Z_{i,j} + Z_{j,k} - 1 &\geq Z_{i,k}, \forall \tau_i,\tau_j,\tau_k \label{eq:opTree6}
\end{align}
Equation \ref{eq:opTree1} states that we want to find the decision tree, $T^*$, among all possible trees, $T$, that minimizes an expected loss function, $L$. Recall from Section \ref{sec:TechnicalApproach} that $y^m_{\tuple{\tau_i,\tau_j}}$ is the binary label given to an observation to indicate whether the human demonstrator took action $\tau_i$ or $\tau_j$. Further, $^{rank}\theta^m_{\tuple{\tau_i,\tau_j}}$ is the corresponding feature vector from that observation. Equations \ref{eq:opTree2} through \ref{eq:opTree5} force the pairwise comparisons to agree with the cumulative ranking. $Z_{i,j}$ is a binary decision variable that is equal to 1 when $\tau_i$ is expected to be chosen over $\tau_j$ and 0 when $\tau_j$ is expected to be chosen over $\tau_i$. Recall that $M$ is a large positive number that allows one to formulate linear, conditional constraints. Finally, Equation \ref{eq:opTree6} requires that the transitive property holds for $T$. Specifically, if $\tau_i$ is predicted to be more likely than $\tau_j$ (i.e, $Z_{i,j} = 1$), and $\tau_j$ (i.e, $Z_{j,k} = 1$) is more likely than $\tau_k$, then $\tau_i$ should also be predicted to be more likely than $\tau_k$ (i.e, $Z_{i,k }= 1$).

\end{document}